\documentclass{article}
\usepackage{arxiv}
\usepackage[utf8]{inputenc} 
\usepackage[T1]{fontenc}    
\usepackage{hyperref}       
\usepackage{url}            
\usepackage{booktabs}       
\usepackage{longtable}
\usepackage{multirow}       
\usepackage{amsfonts}       
\usepackage{nicefrac}       
\usepackage{microtype}      
\usepackage{lipsum}
\usepackage{graphicx}
\usepackage{subcaption}
\usepackage{amsmath}

\graphicspath{ {./images/} }

\title{Large Language Models Struggle with Ethnographic Text Annotation}

\author{
 Leonardo S. Goodall\textsuperscript{†}\\
  Calleva Research Centre\\
  Oxford Internet Institute\\
  University of Oxford, UK\\
  \And
 Dor Shilton\textsuperscript{†}\\
  Cohn Institute for the History and Philosophy of Science and Ideas\\
  Tel Aviv University, Israel\\
  \texttt{dorshilt@mail.tau.ac.il}\\
  \And
 Daniel Austin Mullins\\
  Birkbeck College\\
  University of London, UK\\
  Centre for the Study of Social Cohesion\\ 
  University of Oxford, UK\\
   \And
 Harvey Whitehouse\\
  Centre for the Study of Social Cohesion\\
  University of Oxford, UK\\
}

\begin{document}
\maketitle
\begingroup
\renewcommand\thefootnote{†}
\footnotetext{Both authors contributed equally to this work.}
\endgroup
\begin{abstract}
Large language models (LLMs) have shown promise for automated text annotation, raising hopes that they might accelerate cross-cultural research by extracting structured data from ethnographic texts. We evaluated 7 state-of-the-art LLMs on their ability to annotate 121 ritual features across 567 ethnographic excerpts. Performance was limited, falling well below levels required for reliable automated annotation. Longer texts, features requiring ordinal distinctions, and ambiguous constructs proved particularly difficult. Human inter-coder reliability set an approximate ceiling on LLM accuracy: features that human coders found difficult to agree upon were also difficult for LLMs. Yet even on features where humans reliably agreed, models fell short of human performance. Our findings suggest that LLMs cannot yet substitute for human expertise in ethnographic annotation.
\end{abstract}


\section*{Introduction}
The ethnographic record is one of anthropology's richest contributions to the studies of cultures across the world and throughout history. This corpus, built from immersive fieldwork and rich cultural interpretation, should in principle lend itself well to cross-cultural anthropology: the systematic and comparative study of human cultural variation. Such research typically proceeds by formulating a research question, selecting an appropriate sample of cultures, identifying relevant ethnographic sources, coding those sources according to predefined variables (e.g., text annotation), and conducting statistical analyses on the resulting dataset. However, this process—particularly text annotation—remains methodologically resource-intensive \cite{wang_want_2021} and epistemologically-contested among anthropologists, making such comparative analyses relatively infrequent.

Extracting structured, generalisable data (``thin descriptions") from ethnographic texts (``thick descriptions") is slow, costly, and highly dependent on expert labour, particularly as ethnographic and monographic texts can be lengthy, linguistically obscure, and resistant to abstraction \cite{whitehouse_against_2024}. Moreover, manual annotation necessitates specialists and domain experts fluent in both the ethnographic idioms of the source texts and the comparative frameworks used to code them. The result is a methodological bottleneck: the data needed to test cross-cultural hypotheses exists, but remains buried in hundreds of thousands of pages of prose. Compounding this is a deeper disciplinary hesitance. Sociocultural anthropology, particularly in its post-structural and interpretive turns, has committed itself to ``interpretive exclusivism" (that cultural systems ought to be interpreted and not explained; \cite{whitehouse_against_2024}) and cultural relativism (that societies should be understood on their own terms), hardening into a research orthodoxy that resists comparison and generalisation. Nonetheless, landmark projects such as the electronic Human Relations Area Files (eHRAF) have demonstrated the value of cross-cultural datasets for investigating hunter-gatherer social learning \cite{terashima_cross-cultural_2016}, the recurrence and structure of informal, pre-doctrinal religious practices \cite{boyer_informal_2020}, variation in music-making practices \cite{shilton_group_2023, mehr_universality_2019}, and the cross-cultural stability of beliefs in mystical harm and witchcraft \cite{singh_magic_2018}.


The role of large language models (LLMs) in qualitative research is increasingly being recognised \cite{grossmann_ai_2023, liao_llms_2024}, particularly in relation to data collection and analysis workflows \cite{schroeder_large_2025}. Pre-trained on extensive pools of natural language drawn from books, articles, websites, and other digital sources \cite{brown_language_2020, radford_language_2019}, LLMs have demonstrated exceptional language understanding \cite{brown_language_2020, hendrycks_measuring_2021, rae_scaling_2022, wang_superglue_2019, yang_harnessing_2023}, capturing with it various patterns of cultural knowledge, linguistic idiosyncrasies, common-sense reasoning \cite{liu_generated_2022}, and even human behaviour \cite{park_generative_2023, park_social_2022, horton_large_2023, hewitt_predicting_2024}. For qualitative research, LLMs have been employed to generate initial thematic codes, assist with text annotation, and streamline otherwise time-consuming aspects of manual coding \cite{schroeder_large_2025, gao_collabcoder_2024, gebreegziabher_patat_2023}, whilst dramatically reducing the time and cost of per-annotation labour \cite
{gilardi_chatgpt_2023, grossmann_ai_2023, tornberg_chatgpt-4_2023, 
bang_multitask_2023, qin_is_2023, ziems_can_2024}.


Indeed, when evaluated against human annotators, some LLMs have been shown to match and even exceed human performance in classification and annotation tasks \cite{gilardi_chatgpt_2023, tornberg_chatgpt-4_2023, rathje_gpt_2023, wang_want_2021}, even without any task-specific re-training (i.e., ``zero-shot" learning \cite{qin_is_2023}). Lupo and colleagues \cite{lupo_towards_2024} applied OpenAI's GPT (3.5 and 4) and various open-source models (Llama 2, Mixtral 8x7b, finetuned GPT-SW3) to theoretically complex annotation tasks in political discourse including non-English texts, jargon-rich text, and ambiguous constructs. Notably, whilst human annotators achieved only moderate inter-coder reliability reflecting task complexity, the models showed comparable performance across a range of prompting strategies, with some model performance dropping when handling subtle conceptual distinctions. Dunivin and colleagues \cite{dunivin_scalable_2024} showed that while ChatGPT was effective in identifying general themes across social science domains like gender and inclusion, it struggled with interpretive nuance where implicit information was required. Tai and colleagues \cite{tai_examination_2024} examined ChatGPT-3.5's ability to conduct deductive coding on 3 interview excerpts and 5 codes, finding low to moderate agreement with human annotators.

However, enthusiasm for LLM-assisted qualitative annotation has been tempered by a range of concerns. Importantly for ethnographic contexts, concerns have been noted about the ability of LLMs to handle subtle or culturally nuanced distinctions, particularly in interpretive domains where ambiguity and context are central \cite{dunivin_scalable_2024, dentella_systematic_2023, bisbee_synthetic_2023}. Generally, LLMs are prone to hallucinations, incorrect or biased annotations (especially in non-English texts), anchoring effects (where model outputs subtly shape human judgement during analysis), and ethical and privacy risks of LLM use \cite{schroeder_large_2025, friedman_should_2024, yan_human-ai_2023, santurkar_whose_2023, abdurahman_perils_2023, feuston_putting_2021, jiang_supporting_2021, lam_concept_2024, overney_sensemate_2024}. Moreover, recent large-scale evaluations show that LLM annotations can be unstable across runs, models, and time, exhibit low inter-coder reliability relative to human coding, and materially alter downstream statistical inferences even under tightly controlled prompting conditions \cite{barrie_replication_2025, yang_data_2025}. Related work demonstrates that LLM annotation pipelines are also vulnerable to systematic manipulation, whereby changes in prompts or model selection can be used to recover almost any substantive conclusion from the same data, effectively introducing a powerful and opaque researcher degree of freedom \cite{baumann_large_2025}. More fundamentally, treating LLM outputs as neutral or objective measurements risks fostering an illusion of methodological validity and greater explanatory depth, obscuring the assumptions and abstractions introduced by automated annotation pipelines \cite{messeri_artificial_2024}, as well as pre-training data biases and structural limitations of the models themselves \cite{bommasani_opportunities_2021}.

Critically, there is also tension with core qualitative research values, which emphasise interpretive intimacy with data \cite{feuston_putting_2021, soden_evaluating_2024}, and a broader unease that such tools may reflect or reinforce positivist epistemologies in fields grounded in interpretive traditions \cite{soden_evaluating_2024, baumer_comparing_2017, crabtree_h_2025, nelson_computational_2020}. Indeed, some scholars have cautioned against researcher displacement \cite{feuston_putting_2021} and instead stressed the importance of using LLMs, if necessary, to augment human hermeneutic interpretation \cite{bail_can_2023, dunivin_scalable_2024, ziems_can_2024, dunivin_scaling_2025} and the exploratory, ambiguous nature of qualitative inquiry \cite{jiang_supporting_2021, feuston_putting_2021}. Some scholars, however, frame this shift as part of a broader evolution in qualitative methods, comparable to earlier transitions involving interview recordings, automated transcription, and digital ethnographic methods \cite{brown_going_2002, bryda_qualitative_2023, horst_digital_2012}, all of which faced their own initial contentions \cite{blismas_computer-aided_2003, gilbert_going_2002, spector_tools_2014}. As Schroeder and colleagues \cite{schroeder_large_2025} suggest, such developments may continue to blur the boundary between positivist and interpretivist traditions, opening new methodological spaces without fully collapsing the distinction \cite{lam_concept_2024}.

In summary, the ethnographic record is an unusually rich resource for systematic cross-cultural analysis, yet remains underexploited due to epistemological hesitations and the substantial labour required for annotation. LLMs have shown promise as tools for qualitative coding and may therefore render large-scale comparative analysis more feasible, as first suggested by Dubourg and colleagues \cite{dubourg_step-by-step_2024}. However, ethnographies have traditionally taken as their subjects cultures that differ substantially from Western societies, tending to focus on behaviours that would appear irrational and puzzling to outsiders \cite{gell_art_2013}, often in great detail, and sometimes using obscure language. Because LLMs learn statistical regularities from predominantly contemporary, industrialised, and culturally mainstream corpora, they may systematically misinterpret, flatten, or normalise precisely those features that ethnography seeks to foreground. This article, then, takes the first step in offering a proof-of-concept study that evaluates whether LLMs can reliably annotate ethnographic texts with culturally meaningful features at a scale and speed that could render large-scale cross-cultural comparison more tractable.

\section*{Results}

\begin{table*}[htbp]
\centering
\begin{tabular}{llllll}
\toprule
\textbf{Model} & \textbf{Developer} & \textbf{Parameters} & \textbf{Access} & \textbf{Web-enabled} & \textbf{Reference} \\
\midrule
Qwen 3 Instruct & Alibaba Cloud & 4B & Open-source & No & \cite{yang_qwen3_2025} \\
Llama 3.2 Instruct & Meta & 3B & Open-source & No & \cite{grattafiori_llama_2024} \\
GPT-OSS & OpenAI & 120B (MoE) & Open-source & No & \cite{openai_gpt-oss-120b_2025} \\
DeepSeek V3.1 & DeepSeek & 671B (MoE) & Open-source & No & \cite{deepseek-ai_deepseek-v3_2025} \\
GPT-5 Nano & OpenAI & -- & Proprietary & No & \cite{openai_gpt-5_2025} \\
Claude Sonnet 4.5 & Anthropic & -- & Proprietary & No & \cite{anthropic_system_2025} \\
Perplexity Sonar & Perplexity & -- & Proprietary & Yes & \\
\bottomrule
\end{tabular}
\caption{\textbf{Summary of large language models evaluated.} \textit{Note.} Parameter counts for proprietary models are undisclosed.}
\label{tab:models}
\end{table*}

We evaluated 7 large language models (LLMs; Figure \ref{tab:models}) on their ability to annotate two complementary datasets, both derived from the same corpus of 567 ethnographic texts describing rituals in a globally representative sample of 73 unique cultures. The first dataset comprises 115 ritual features such as ritual function, duration, and various actions performed during the ritual (``morphospace dataset" hereafter). Annotated by a single expert coder (except two features, see Methods), this dataset enables a broad assessment of LLM performance across diverse feature types and categories. The second dataset is a previously unpublished one focused on 6 types of synchronous behaviour (``synchrony dataset" hereafter). With two annotations per feature, this dataset enables us to quantify human inter-coder reliability, assess whether task difficulty constrains LLM performance, and compare error patterns between human and LLM annotators. We further examine what factors predict LLM accuracy using mixed-effects models that incorporate model, text, and task characteristics.

\subsection*{LLMs show poor annotation performance for ethnographic texts}

\begin{figure*}
    \centering
    \includegraphics[width=1\linewidth]{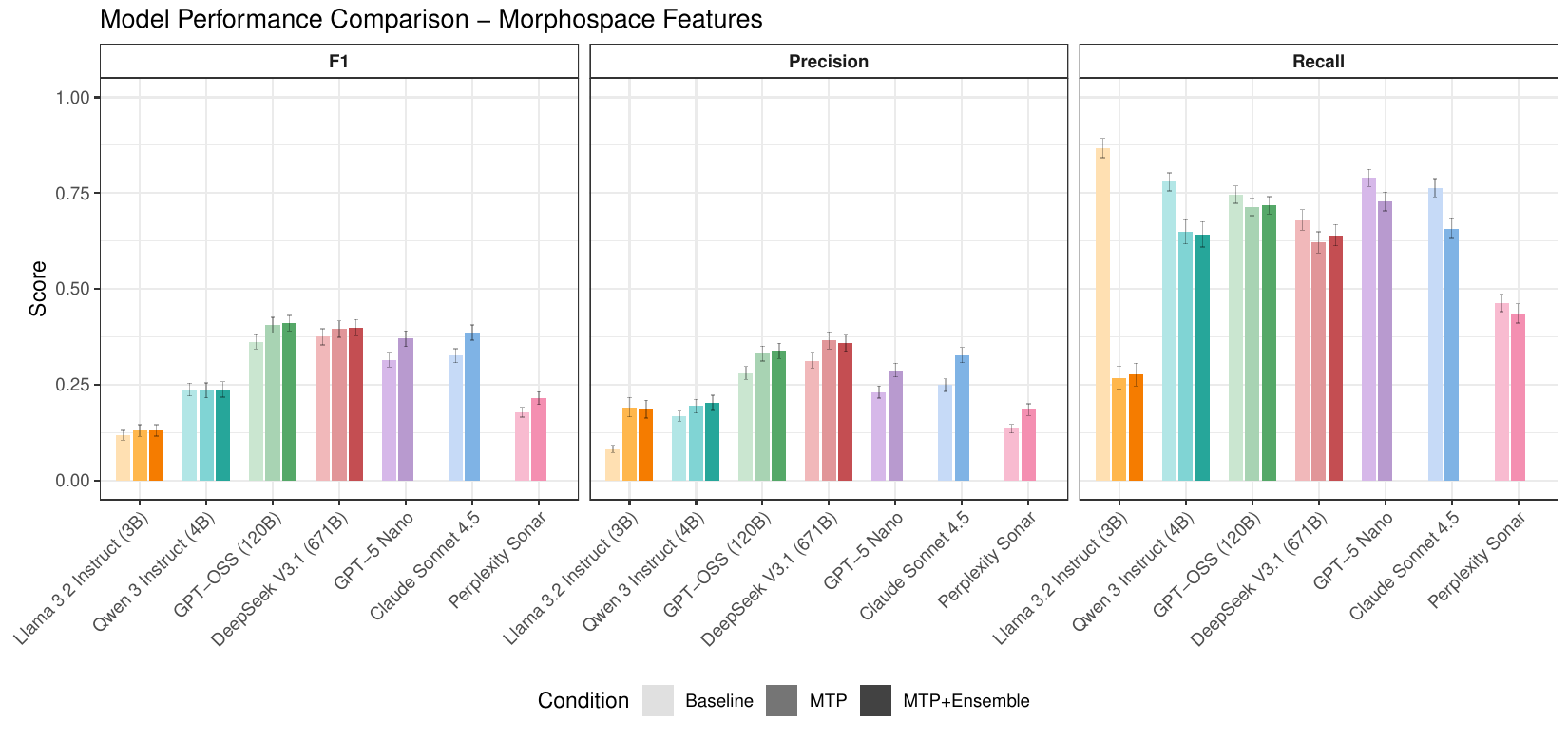}
    \caption{\textbf{Prompting strategy and ensembling improve LLM annotation performance across 115 morphospace features.} Mean F1 score for each model under three conditions: baseline (zero-shot prompting), MTP (multi-task prompting), and MTP+Ensemble (majority voting across MTP runs). Error bars indicate 95\% confidence intervals. MTP consistently outperforms baseline prompting, with ensembling providing additional gains for most models. Due to cost restrictions, only open-source models included ensembling.}
    \label{fig:model-performance-bar-all}
\end{figure*}

Across all models and prompting conditions, comparing LLM outputs to expert human annotations in the morphospace dataset, performance was consistently poor with mean F1 scores ranging from 0.12 to 0.41 (Figure \ref{fig:model-performance-bar-all}). The best-performing configuration was DeepSeek V3.1 with multi-task prompting and ensemble sampling (F1=0.41), followed closely by GPT-OSS 120B under the same condition (F1=0.41). Proprietary models such as GPT-5 Nano and Claude Sonnet 4.5 achieved comparable performance (F1=0.37 and 0.39, respectively). The smallest open-source models, Llama 3.2 Instruct (3B) and Qwen 3 Instruct (4B), performed worst, with F1 scores below 0.25. Interestingly, the Llama model often failed to comply with multi-task instructions giving invalid model inferences. The web-enabled Perplexity Sonar, despite access to external knowledge, showed no advantage over offline models (F1=0.23).

Prompting strategy had modest effects on performance. Compared to the zero-shot baseline (annotating one feature per ritual at a time), multi-task prompting (annotate all features in a predefined category at once) improved F1 scores by 0.02 to 0.07 points for most models, with DeepSeek V3.1 showing the largest gain. Ensemble sampling (taking the mode of 10 repetitions) provided additional but diminishing returns, typically adding 0.01 to 0.02 points over multi-task prompting alone. Notably, the relationship between precision and recall varied across models: smaller models such as Llama 3.2 and Qwen 3 exhibited high recall but low precision, suggesting a tendency to over-predict feature presence, whereas larger models achieved more balanced trade-offs.

\begin{figure*}
    \centering
    \begin{subfigure}[t]{0.49\linewidth}
        \centering
        \includegraphics[width=\linewidth]{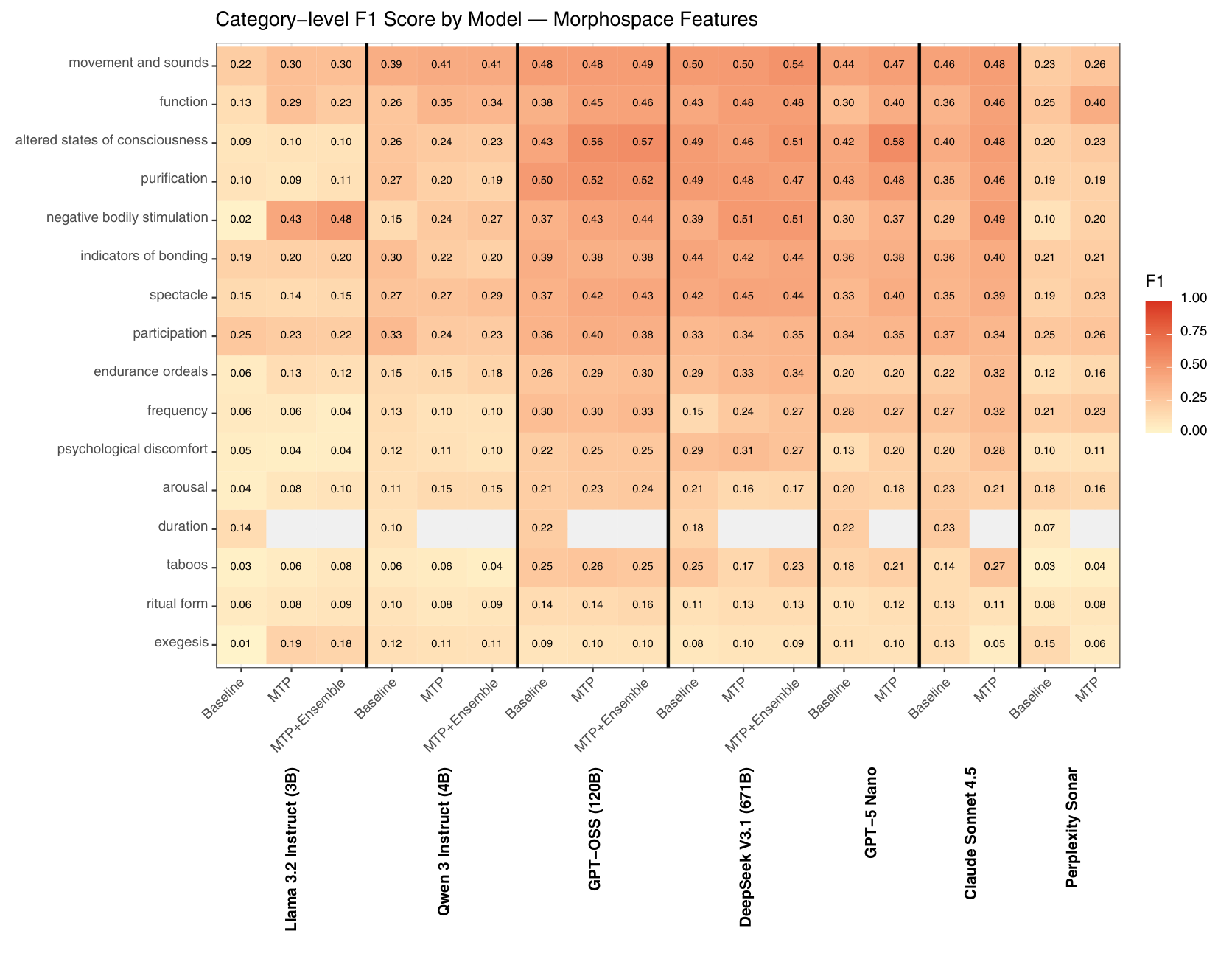}
        \caption{}
        \label{fig:model-performance-category-all}
    \end{subfigure}\hfill
    \begin{subfigure}[t]{0.49\linewidth}
        \centering
        \includegraphics[width=\linewidth]{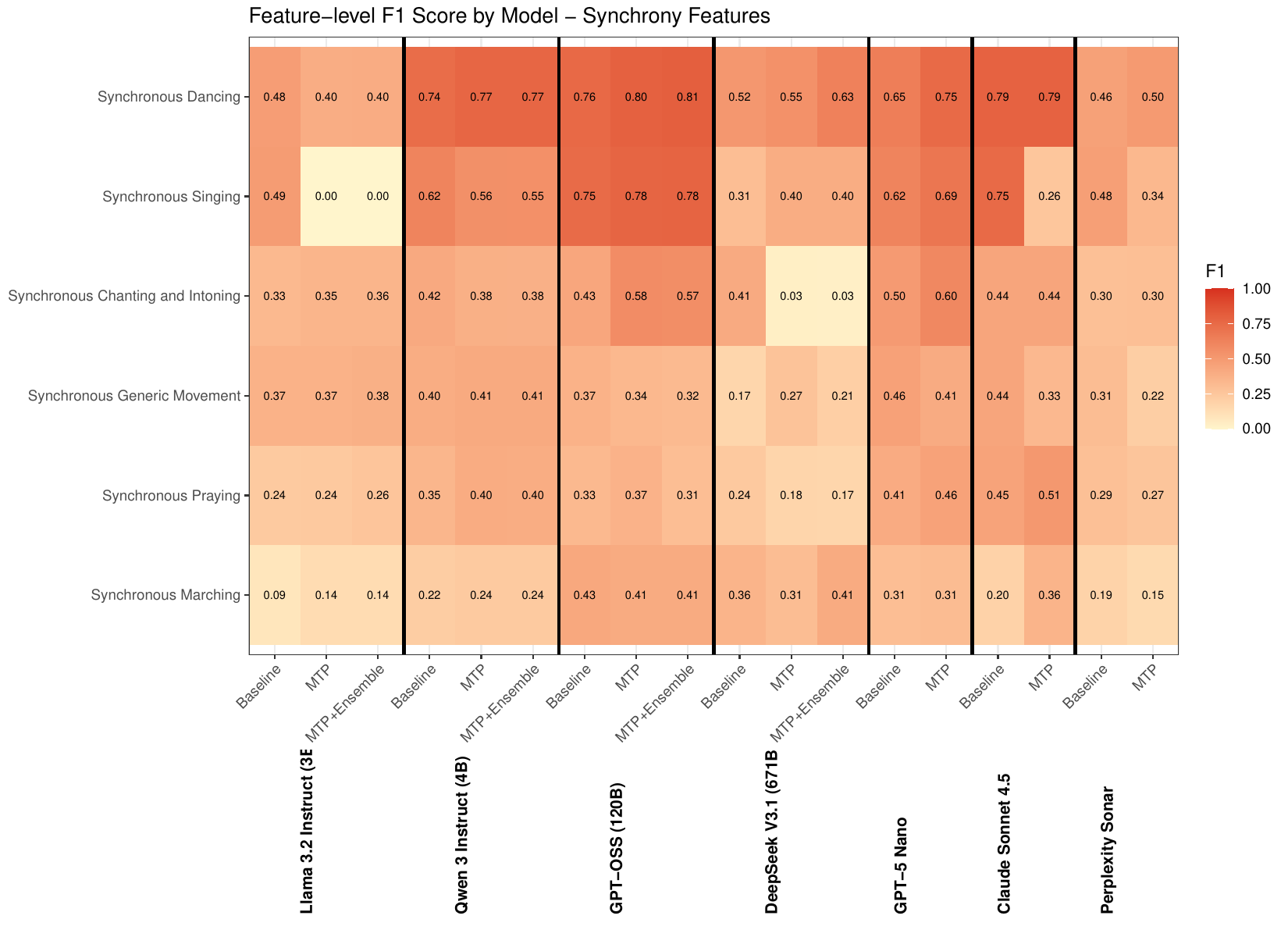}
        \caption{}
        \label{fig:model-performance-feature-synchrony}
    \end{subfigure}
    \caption{\textbf{LLM performance across morphospace categories and synchrony features.} (a) Heatmap showing mean F1 score for each model–condition combination (columns) across feature categories (rows). Categories are ordered by average performance. Duration, comprising a single feature, was evaluated under baseline conditions only as multi-turn prompting requires multi-feature iteration. Performance is highest for well-defined categorical features and lowest for subjective, context-dependent attributes and multi-option features. (b) Feature-level F1 scores for synchrony feature annotations across models and conditions. Features requiring identification of unambiguous coordinated actions (e.g., dancing and singing) show higher performance than those involving nuanced judgements about group coordination (e.g., generic synchronous movement). This pattern suggests that feature complexity partly drives classification difficulty.}
    \label{fig:model-performance-combined}
\end{figure*}

Performance varied substantially across feature categories (Figure \ref{fig:model-performance-category-all}). Features relating to ritual function (e.g., funerary, initiation, newborn ceremonies) and movement (e.g., dancing, singing) were annotated with relatively higher accuracy (F1 $>$ 0.60 for the best models), likely because these features are explicitly described in ethnographic texts. In contrast, features requiring interpretive inference, such as psychological discomfort, arousal levels, and ritual form, proved considerably more difficult (F1 $<$ 0.30). Binary features were generally easier to classify than multi-class features, which required distinguishing among multiple ordinal or categorical options.

For the synchrony dataset comprising 6 features (synchronous singing, chanting, praying, marching, dancing, and generic movement) performance was marginally higher but still limited
(Figure \ref{fig:model-performance-feature-synchrony}). The best-performing model on this dataset was GPT-5 Nano with multi-task prompting (F1=0.54), followed by GPT-OSS 120B (F1=0.54). Individual feature analysis revealed that synchronous dancing and singing were the easiest to detect (F1 $>$ 0.78), while synchronous marching proved most difficult (F1=0.43).



All models across both datasets outperformed a majority-class baseline classifier, which always predicts the most frequent class for each feature. Although, given the severe class imbalance (most features are absent in most rituals), this baseline achieves near-zero F1 for the vast majority of features. The mean baseline F1 across all features for the first, larger dataset was 0.02, compared to 0.47 for the best LLM configuration. While this confirms that LLMs may capture meaningful signal beyond class frequency, the absolute performance remains far below levels required for reliable automated annotation.

\subsection*{Human agreement sets a ceiling on LLM performance}

The synchrony dataset, where two human coders independently annotated each ritual, allowed us to examine whether human inter-coder reliability constrained LLM performance. Human coders achieved moderate overall agreement, with a mean Cohen's kappa of 0.57 and mean raw agreement of 89\% across the six features.
However, reliability varied substantially by feature. Synchronous singing showed near-perfect agreement ($\kappa$=0.92), while synchronous generic movement proved difficult even for humans ($\kappa$=0.25). Only one feature achieved the conventional threshold for substantial agreement ($\kappa \geq 0.80$), and two fell below fair agreement ($\kappa < 0.40$).

\begin{figure*}
    \centering
    \includegraphics[width=1\linewidth]{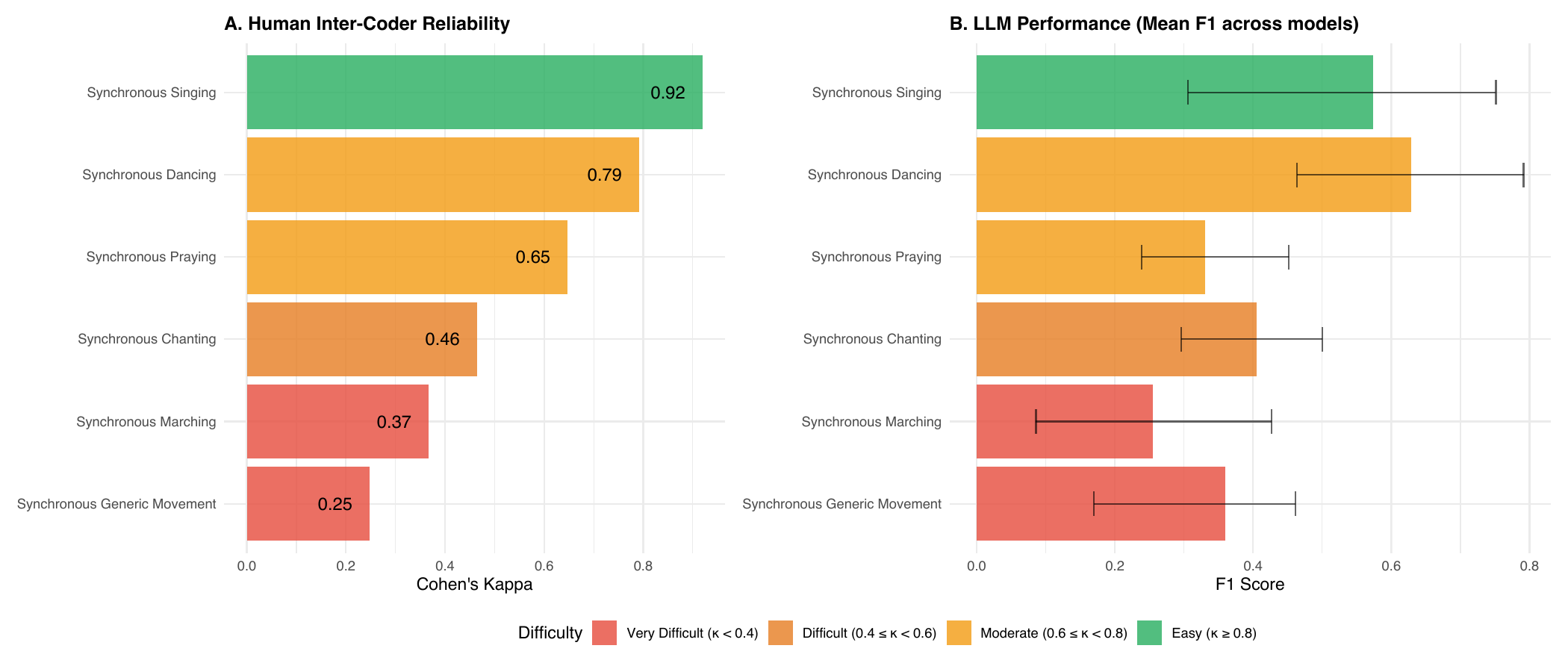}
    \caption{\textbf{Human inter-coder reliability predicts LLM performance.} (a) Cohen's $\kappa$ measuring agreement between two independent human coders for each synchrony feature. (b) Mean F1 score across all LLM models for the same features; error bars show the range across models. Colour indicates difficulty category based on human agreement: features that humans find difficult to code reliably (low $\kappa$) also prove challenging for LLMs, suggesting inherent ambiguity in the ethnographic texts and/or task rather than model-specific limitations.}
    \label{fig:icr-llm-performance}
\end{figure*}

Human reliability was predictive of LLM performance. Across all models, we observed a positive correlation between human inter-coder kappa and LLM F1 score ($r$=0.61). Features that humans found easier to agree upon were also easier for LLMs to classify correctly. For the single feature with substantial human agreement (singing, $\kappa$=0.92), LLMs achieved a mean F1 of 0.57; for the two features with poor human agreement ($\kappa < 0.40$), mean LLM F1 dropped to 0.31. This pattern suggests that task difficulty, as indexed by human disagreement, sets an approximate ceiling on automated annotation performance (Figure \ref{fig:icr-llm-performance}).

Importantly, LLMs did not fully reach the human reliability ceiling for the easier features. For synchronous singing, where humans achieved near-perfect agreement ($\kappa$=0.92), the best-performing LLM (Claude Sonnet 4.5) reached F1=0.75. For synchronous dancing ($\kappa$=0.79), however, the best LLM matched human reliability (F1=0.79), demonstrating that parity is achievable under favourable conditions. For the most difficult features (marching, generic movement), some LLMs slightly exceeded human kappa, suggesting that when annotation is inherently ambiguous, automated approaches may perform comparably.

Error pattern analysis revealed systematic differences between human and LLM annotators. Human coders exhibited low false positive rates (mean 4\%) but higher false negative rates (mean 51\%), indicating a conservative bias toward under-reporting feature presence. LLMs showed the opposite tendency: most models exhibited elevated false positive rates, with smaller models such as Llama 3.2 Instruct predicting presence in nearly all cases (FP rate $>$ 98\%). The best-performing models (GPT-OSS 120B, Claude Sonnet 4.5) achieved more balanced error profiles, though their false positive rates still exceeded those of human coders. Consistent with these divergent error patterns, LLMs agreed with each other far less than humans did (mean LLM-LLM $\kappa$=0.23 vs.\ Human-Human $\kappa$=0.57), suggesting that different models interpret ambiguous ethnographic descriptions in systematically different ways, limiting the potential for simple ensemble approaches to improve reliability.

\subsection*{Feature complexity and text length drive errors}

To identify what drives variation in LLM accuracy, we fitted nested mixed-effects models at two levels of analysis: ritual-level (using F1 score as the outcome) and prediction-level (using binary correctness for each feature annotation). Predictors were grouped into three classes: model factors (model identity, prompting condition), text factors (character length, readability, lexical diversity, geographic region, publication year), and task factors (feature type, description length, base rate). This analysis was conducted for the morphospace dataset only.

At the ritual level, model factors alone explained 36.5\% of variance in F1 scores (marginal $R^2$). Adding text factors increased this modestly to 38.6\%, and task factors contributed a further 0.9\% (Table~\ref{tab:diagnostics_r2}). Including random intercepts for ritual raised the conditional $R^2$ to 71.6\%, indicating that much of the explainable variance is attributable to ritual-level heterogeneity rather than to measured text or task characteristics, with substantial residual variation remaining both within and between rituals. Variance decomposition showed that 53.2\% of residual variance lay between rituals, with 46.8\% within rituals across model-condition combinations. Text length was negatively associated with performance ($\beta$=-0.015, $p<$.001): longer ethnographic excerpts yielded lower F1 scores. Neither readability nor lexical diversity significantly predicted accuracy, despite considerable variation across texts: the Gunning Fog index ranged from 6.4 to 19.9 (M=12.0, SD=2.1), indicating reading levels from elementary school to graduate level, while the Measure of Textual Lexical Diversity (MTLD; where higher values indicate greater word diversity) ranged from 22.5 to 134.8 (M=65.8, SD=16.7), spanning from repetitive to highly varied prose. Among task factors, longer feature descriptions were associated with poorer performance ($\beta$=-0.024, $p<$.001). 

Geographic region showed modest effects: texts from Oceania exhibited significantly lower F1 scores than other regions ($\beta$=-0.043, $p$=.003), with North America showing a similar but non-significant trend ($\beta$=-0.030, $p$=.069). Separate Kruskal-Wallis tests for each model confirmed regional variation, with both North America and Oceania showing significantly lower F1 scores for most models (see Supplementary Materials). A notable exception was Perplexity Sonar, the only web-enabled model evaluated: it performed significantly better on European texts than on texts from all other regions.

\begin{table}[htbp]
\centering
\begin{tabular}{lcc}
\toprule
\textbf{Model} & \textbf{Marginal $R^2$} & \textbf{Conditional $R^2$} \\
\midrule
\multicolumn{3}{l}{\textit{Ritual level (F1 score)}} \\
M1: Model factors & 0.36 & 0.71 \\
M2: + Text factors & 0.39 & 0.72 \\
M3: + Task factors & 0.39 & 0.72 \\
\midrule
\multicolumn{3}{l}{\textit{Prediction level: Positive stratum (detection)}} \\
M1: Model factors & 0.04 & 0.30 \\
M2: + Text factors & 0.04 & 0.30 \\
M3: + Task factors & 0.22 & 0.37 \\
\midrule
\multicolumn{3}{l}{\textit{Prediction level: Negative stratum (specificity)}} \\
M1: Model factors & 0.25 & 0.55 \\
M2: + Text factors & 0.29 & 0.55 \\
M3: + Task factors & 0.35 & 0.52 \\
\bottomrule
\end{tabular}
\caption{Variance explained by nested mixed-effects models. Marginal $R^2$ reflects fixed effects only; conditional $R^2$ includes random effects.}
\label{tab:diagnostics_r2}
\end{table}

At the prediction level, we stratified observations by ground truth class to examine detection (correct identification of present features; 15.6\% of cases) and specificity (correct rejection of absent features; 84.4\% of cases) separately. This stratification revealed divergent patterns. For detection, model and text factors explained little variance (marginal $R^2$=3.8\% after adding text), but task factors dramatically increased explanatory power to 22.5\%. The dominant predictor was feature type: multiclass features had 90\% lower odds of correct detection than binary features (OR=0.10, 95\% CI: 0.03--0.35), reflecting the difficulty of distinguishing among multiple ordinal or categorical options. Longer texts modestly improved detection (OR=1.08, 95\% CI: 1.05--1.12), perhaps because more extensive descriptions provide additional cues for identifying present features.

For specificity, model factors alone explained a substantial 24.9\% of variance, reflecting large differences in false positive rates across models. Text length was the strongest text-level predictor: each standard deviation increase in log-transformed character length reduced the odds of correct rejection by 42\% (OR=0.58, 95\% CI: 0.56--0.60). This suggests that longer texts generate more opportunities for spurious pattern matching, leading models to incorrectly predict feature presence. Feature base rate was also highly predictive: rarer features (lower base rates) were easier to correctly reject (OR=0.23, 95\% CI: 0.16--0.35). Multiclass features, conversely, were easier to correctly reject than binary features (OR=3.99, 95\% CI: 1.39--11.4), likely because LLMs can partially succeed by predicting the ``absent" or ``zero" category.

The interaction between model and prompting condition revealed heterogeneous effects. For the smallest model (Llama 3.2 Instruct), multi-task prompting produced a paradoxical pattern: detection worsened substantially (OR=0.32 for MTP, OR=0.26 for ensemble), while specificity improved dramatically (OR=90.7 for MTP). This indicates that multi-task prompting shifted the model toward conservative predictions, reducing both true positives and false positives. Larger models showed more modest and balanced effects of prompting strategy.

Finally, we examined whether model certainty in the ensemble+MTP condition predicted annotation accuracy. Certainty was operationalised as the percentage of the 10 ensemble repetitions that agreed with the modal (final) annotation; higher values indicate greater consistency across runs. Across 214,529 predictions from four models, certainty correlated positively with correctness ($r$=0.12, 95\% CI: 0.12--0.12, $p<$.001). Annotations made with 91--100\% certainty achieved 81\% accuracy, compared to 68\% for annotations with 10--50\% certainty. The relationship was strongest for the better-performing models: GPT-OSS 120B ($r$=0.36) and DeepSeek V3.1 ($r$=0.23), but absent for Llama 3.2 Instruct ($r$=-0.01). These findings suggest that ensemble consistency provides a useful proxy for annotation quality, particularly for larger models, and could inform quality control procedures in applied settings.

\section*{Discussion}

This study evaluated whether LLMs can reliably annotate ethnographic texts with culturally meaningful features. The answer, based on our evidence, is a qualified no. Across 7 models, 3 prompting strategies, and 115 ritual features, the best-performing configuration achieved an F1 score of only 0.41. Performance was marginally better on the simpler synchrony dataset but remained far below levels required for unsupervised annotation. These findings suggest that, at present, LLMs cannot substitute for human expertise in extracting structured data from ethnographic prose.

Our diagnostics point to several interrelated factors for why LLMs struggled with this task. First, ethnographic texts are long, dense, and stylistically heterogeneous. Text length was consistently associated with poorer performance: longer excerpts reduced F1 scores at the ritual level and dramatically increased false positive rates at the prediction level. This pattern suggests that LLMs are prone to spurious pattern matching when processing extended prose, finding superficial textual cues that do not correspond to genuine feature presence. The median text in our corpus exceeded 650 words, well beyond the short passages (tweets, headlines, brief survey responses) on which LLMs have demonstrated strong annotation performance in prior studies \cite{gilardi_chatgpt_2023, tornberg_chatgpt-4_2023}.

Second, many ritual features require interpretive inference rather than surface recognition. Features such as ``psychological discomfort," ``level of arousal," and ``ritual form" cannot be identified from explicit textual markers alone; they demand the kind of contextual reasoning and cultural knowledge that human coders bring to the task. Our results showed that multiclass features, which require distinguishing among ordinal or categorical options, were 90\% less likely to be correctly detected than binary features. This suggests that LLMs perform best when the annotation task reduces to presence-versus-absence judgments for concrete, explicitly described phenomena. When the task requires graded assessments or implicit interpretation, performance collapses.

Third, human inter-coder reliability sets an approximate ceiling on LLM performance. Features that human coders found difficult to agree upon were also difficult for LLMs. For the single feature with near-perfect human agreement (synchronous singing, $\kappa$=0.92), LLMs approached but did not match human reliability. For features with poor human agreement ($\kappa < 0.40$), LLMs performed comparably to chance. This correlation implies that some annotation tasks are inherently ambiguous, and no amount of model improvement will resolve disagreements that arise from genuine interpretive indeterminacy. Indeed, the correlation between model certainty and accuracy accentuated this. However, it also suggests that LLMs underperform even where humans can reliably agree, indicating room for improvement on well-defined tasks.

These findings diverge from the optimistic conclusions of prior work evaluating LLMs on text annotation. Studies reporting near-human or superhuman performance have typically used short texts, narrow coding schemes, and domains where ground truth is relatively unambiguous \cite{gilardi_chatgpt_2023, tornberg_chatgpt-4_2023, rathje_gpt_2023}. Ethnographic annotation differs in kind: the texts are longer, the features more numerous and conceptually varied, and the ground truth itself often contested. Our results align with recent cautionary findings showing that LLM annotations can be unstable, exhibit low reliability, and alter downstream inferences \cite{barrie_replication_2025, yang_data_2025, baumann_large_2025}. They also echo concerns that LLMs struggle with interpretive nuance and implicit cultural knowledge \cite{dunivin_scalable_2024, dentella_systematic_2023}.

Several factors beyond those examined here may also contribute to poor performance. LLMs are trained predominantly on Western, English-language corpora, and ethnographic texts describing non-Western rituals may fall outside the distribution of cultural knowledge encoded in model weights. This asymmetric access to cultural knowledge was strikingly illustrated by the web-enabled model, Perplexity Sonar, performing significantly better on European texts than on texts from all other regions, suggesting that its retrieval-augmented architecture draws on online sources that provide richer contextual information for European ethnographic contexts. The archaic or specialised vocabulary common in older ethnographic sources may further compound this gap. Additionally, understanding ritual behaviour often requires background knowledge about cosmology, social structure, and symbolic meaning that is rarely made explicit in the source texts and may not be recoverable from surface-level language patterns alone. Notably, our regional analyses revealed that most models performed significantly worse on texts from North America and Oceania compared to other regions, though the effect sizes were small. Whether this reflects differences in ethnographic writing style, ritual complexity, or cultural distance from training data remains unclear.

Several limitations temper these conclusions. First, we relied on zero-shot and multi-task prompting without extensive prompt engineering or the use of structured codebooks with detailed examples. Prior work suggests that well-designed codebooks with illustrative cases can substantially improve LLM annotation quality \cite{lupo_towards_2024, tai_examination_2024, dunivin_scaling_2025}. Developing such codebooks for 115 features was beyond our capacity, but future work should explore whether codebook-based prompting narrows the gap with human performance. Second, we did not fine-tune any models on our specific task. Although fine-tuning requires technical expertise and computational resources often unavailable to social scientists, it has been shown to improve performance on domain-specific annotation tasks \cite{alizadeh_open-source_2025}. Third, our evaluation used a single resolved ground truth for each feature; alternative approaches such as soft labels or probabilistic annotations might better capture the inherent uncertainty in ethnographic coding. Fourth, whilst multi-task prompting modestly improved performance relative to the zero-shot baseline, likely because annotating multiple features simultaneously allowed the model to leverage shared contextual information across related features, other prompting strategies may prove more effective \cite{sahoo_systematic_2025}.

The ethnographic record should not only be a source of interpretive insight, but also a substrate from which testable hypotheses can be drawn, coded, and compared \cite{whitehouse_against_2024}. The main challenge of cross-cultural research is to produce scientific insights from hundreds of thousands of pages of ethnography. As mentioned, two persistent obstacles beget this impasse: the intensive labour required to extract structured data from ethnographic texts, and a disciplinary resistance to abstraction rooted in anthropological epistemology. Addressing the methodological bottleneck, LLMs have demonstrated potential to significantly speed up and accurately emulate the process qualitative coding and annotation tasks in other domains where, prima facie, it would seem that such promises would extend to the sociocultural, interpretive domain. We find, contrary to expectations, that LLMs instead perform poorly. Researchers hoping to use LLMs for ethnographic coding should proceed with caution, validating outputs against human annotations and reserving automated approaches for features that are concrete, binary, and explicitly described in the source texts. For features requiring interpretation, inference, or cultural knowledge, human expertise remains indispensable. Indeed, these findings highlight a tension at the heart of computational approaches to qualitative data. LLMs may not yet be able to bring contextual and cultural knowledge to bear on ambiguous texts, fully grasp the nuances of complex tasks on long, dense texts. Until LLMs can reliably perform this kind of interpretive work, the promise of automated thick-to-thin data conversion for cross-cultural analysis will remain unfulfilled.

\section*{Methods}

\paragraph{Data} We derived our ethnographic corpus from the Ritual Morphospace Dataset \cite{atkinson_cultural_2011, kapitany_ritual_2020} collected from the electronic Human Relations Area Files (eHRAF) database. After exclusions (N=102 due to missing texts and unclear ritual specifications), we sourced 567 ethnographic texts (median=653 words, range=43–8,924 words) from 73 unique cultures. Figures, illustrations, footnotes (including their in-text numbering), captions, and stylistic formatting (bold, italics, underlining) were omitted. Chapter and (sub)section titles were kept. 

\paragraph{Ritual features} We constructed two datasets of ritual features. Our first (``Morphospace dataset") comprises 115 feature variables similarly taken from eHRAF, grouped into 15 categories: function (15), participation (9), frequency (2), movement and sounds (11), purification (7), altered states of consciousness (7), negative bodily stimulation (13), endurance ordeals (5), psychological discomfort (4), spectacle (23), indicators of bonding (3), taboos (4), arousal (6), ritual form (2), exegesis (3). (Duration of the ritual was a single-item feature and thus excluded from category analyses.) This data was originally used to test the modes of religiosity theory \cite{atkinson_cultural_2011}. All variables were annotated by a single expert coder, except for two features (peak euphoric or dysphoric arousal), which were averaged across 3 ratings provided by 3 annotators (the rounded average was taken as the final annotation).

A novel ``synchrony dataset" comprised six binary feature variables capturing ritual synchrony (singing, chanting, praying, marching, dancing, and generic movement). Each ritual description received two independent annotations, produced by a pool of three expert coders working in rotating dyads. The corpus was divided into three contiguous segments, each assigned to a different dyad, such that across the full dataset all coders contributed while no ritual was coded by more than two individuals. Within each dyad, coders first annotated independently and subsequently resolved any disagreements through discussion, yielding a single agreed coding per ritual. Coders also independently annotated the sex composition of participants, but this variable is not analysed here (see SM, Table \ref{tab:ritual_sync}).

Of the 115 morphospace features, 100 were binary (annotating whether a ritual feature was present or absent) and 15 were multi-class (requiring selection among multiple ordinal or categorical options, such as arousal levels rated 0–5). All synchrony features were binary coded. For the few ritual features in which ``Not Reported" was annotated by the expert coder, we maintained this as an option in the LLM prompt.

\paragraph{Large language models} We evaluated the performance of 7 large language models (LLMs) on the task of annotating the ritual features (Table~\ref{tab:models}). Two classes of models were evaluated: open-source and proprietary. Whilst open-source LLMs offer compelling advantages, including cost-effectiveness, methodological transparency, replicability, and stringent data protection standards \cite{liesenfeld_opening_2023, ollion_dangers_2024, spirling_why_2023}, proprietary models have demonstrated leading performance across various benchmarks. We also included one web-enabled model (Perplexity Sonar), which allowed for the integration of the web as an external knowledge source.

\paragraph{Annotation tasks} All models were evaluated using the same annotation tasks. Models were instructed to annotate a given ethnographic excerpt according to a given ritual feature(s). The prompt included the feature name(s), feature definition(s), and available options.

Where costs allowed, we used 3 prompting strategies to assess model performance: (1) a baseline, single-feature prompt (i.e., one ritual feature and one ethnographic excerpt; ``zero-shot"); (2) multi-task prompt (MTP; i.e., all ritual features belonging to one category and one ethnographic excerpt; e.g., \cite{gozzi_comparative_2024}), and (3) multi-task prompt with ensemble sampling (MTP+Ensemble; i.e., the MTP prompt repeated 10 times with the mode taken as the final annotation). (See \cite{sahoo_systematic_2025} for an extensive survey on prompting strategies.) This applied to all models except the web-enabled Perplexity Sonar. Here, the model was provided a prompt that only included the relevant ritual name, culture, author, date, and ethnography title.

Given limited labelled data and to prioritise usability for non-technical researchers, we adopt prompting-based strategies rather than fine-tuning the models' internal parameters, despite the latter's known performance benefits \cite{alizadeh_open-source_2025}. Indeed, appropriate fine-tuning requires machine learning expertise and infrastructure often inaccessible to many social scientists, whereas prompting allows intuitive, low-barrier task definition using natural language. 

\paragraph{Evaluation}
We conducted a series of analyses to assess LLM performance and diagnose sources of error. These comprised the following: (1) performance comparison across models and prompting conditions against human ground truth annotations; (2) reliability analysis, computing Cohen's kappa and raw agreement between human coders in the synchrony dataset and comparing LLM-human agreement to this benchmark to examine whether human inter-coder reliability sets a ceiling on LLM performance; and (3) performance diagnostics using mixed-effects regression models to identify which model, text, and task characteristics predict classification accuracy at both the ritual and prediction levels.

Firstly, we computed precision, recall, and F1 scores for each model and prompting condition by comparing LLM outputs to human ground truth annotations. Precision measures the proportion of features that the model predicted as present that were genuinely present according to human coding; it captures how often the model is correct when it makes a positive prediction. Recall measures the proportion of genuinely present features that the model successfully detected; it captures how completely the model identifies what is actually there. F1 is the harmonic mean of precision and recall, providing a single summary measure that balances both concerns and ranges from 0 (complete failure) to 1 (perfect agreement). When a model never predicted a particular class (yielding undefined precision), we treated that class as having zero precision, recall, and F1; an F1 of 0 thus indicates complete failure to detect that feature or class. For binary features, we calculated these metrics treating presence as the positive class. For multi-class features, we computed macro-averaged metrics across all classes present in the ground truth. Four features were excluded due to annotation inconsistencies. To contextualise model performance, we also computed the F1 score of a majority-class baseline classifier that always predicts the most frequent class, providing a lower bound against which to evaluate whether LLMs offer meaningful predictive value (See Supplementary Materials).

Secondly, for the synchrony dataset where dual independent human annotations were available, we computed Cohen's kappa and raw agreement between the two human coders to establish ground truth reliability. We then calculated Cohen's kappa between each LLM and the resolved human annotations, comparing this to the human-human benchmark to assess whether inter-coder reliability sets a ceiling on LLM performance. We further examined LLM–LLM agreement and whether LLMs exhibited similar error patterns to human coders by comparing false positive and false negative rates across annotators.

Finally, to diagnose sources of performance variation, we fitted nested mixed-effects models examining three classes of predictors: model factors (model identity, prompting condition, and their interaction), text factors (character length, readability, lexical diversity, geographic region, and publication year), and task factors (feature type, description length, and base rate). Readability was measured using the Gunning Fog index, which estimates the years of formal education required to understand a text on first reading; higher values indicate more complex prose \cite{gunning_fog_1969}. Lexical diversity was measured using the Measure of Textual Lexical Diversity (MTLD), which quantifies vocabulary richness by assessing how long a text can maintain a given level of word variation \cite{mccarthy_mtld_2010}; higher values indicate greater vocabulary diversity. At the ritual level, we modelled F1 score as a continuous outcome using linear mixed-effects regression with random intercepts for ritual. At the prediction level, we modelled binary correctness using generalised linear mixed-effects models with random intercepts for ritual and feature. Given severe class imbalance (84.6\% of ground truth labels were absent), we stratified predictions into positive and negative strata to separately assess detection and specificity. Fixed effects were added incrementally across four nested models; we report marginal and conditional $R^2$, likelihood ratio tests, and odds ratios with 95\% confidence intervals. Geographic variation in performance using Kruskal-Wallis tests across world regions was tested. Finally, we tested for correlation between model certainty (taken as the percentage of the LLM's modal annotation in the ensemble condition) and accuracy. 

\section*{Acknowledgements}
We thank A. Giritlioglu for initial discussions on this paper and J.-L. Tucker, B. Tunçgenç, M. Newson, and Q. Atkinson for their work on the synchrony dataset.

\section*{Author contributions} L.S.G. and D.S. contributed equally and jointly led the conceptualisation of the study. L.S.G. developed and implemented the analysis pipeline, conducted formal analysis and validation, generated visualisations, and co-wrote the original draft of the manuscript. D.S. contributed to methodology, co-wrote the original draft, and oversaw project administration and resources. D.A.M. was responsible for data curation and provided essential resources. H.W. provided supervision, contributed to writing (review and editing), and supported the project with key resources.

\section*{Data availability}
Model predictions and human annotations are available upon reasonable request. Raw ethnographic texts are sourced from the eHRAF World Cultures database (\href{https://ehrafworldcultures.yale.edu/}{https://ehrafworldcultures.yale.edu/}), which requires an institutional licence; we therefore cannot redistribute these primary sources.

\section*{Code availability}
Analysis code is available upon reasonable request.

\section*{Supplementary materials}

\subsection*{Synchrony Features}

All synchrony features, their descriptions, and annotation options given both to human coders and LLMs is provided in Table \ref{tab:ritual_sync}.

\begin{table*}[h]
    \centering
    \begin{tabular}{p{4cm} p{9cm} p{3cm}}
    \toprule
    \textbf{Feature Name} & \textbf{Feature Description} & \textbf{Feature Options} \\
    \midrule
    Synchronous Singing & two or more people singing the same melody and rhythm in unison & 0=absent, 1=present \\
    Synchronous Chanting and Intoning & two or more people vocalizing rhythmically, repetitively, and simultaneously with little or no rise and fall of the pitch of the voice & 0=absent, 1=present \\
    Synchronous Praying & vocalized requests of thanks or praise to a deity or other object of worship in a rhythmic and repetitive manner by two or more people at the same time & 0=absent, 1=present \\
    Synchronous Marching & two or more people walking or proceeding in unison with a regular measured step & 0=absent, 1=present \\
    Synchronous Dancing & two or more people leaping, skipping, hopping, or gliding with measured steps and rhythmical movement of the body in unison & 0=absent, 1=present \\
    Synchronous Generic Movement & rhythmic and repetitious bodily movements other than marching and dancing by two or more people in unison & 0=absent, 1=present \\
    \bottomrule
    \end{tabular}
    \caption{Ritual synchrony features, descriptions, and annotation options.}
    \label{tab:ritual_sync}
\end{table*}

\subsection*{Prompts}
All prompts passed to LLMs are given below.

\paragraph{Baseline prompt}
\begin{quote}
    \small
    \raggedright
    \textbf{Instructions}\\
    You are given an ethnographic excerpt to classify.\\
    - Carefully read the excerpt in full.\\
    - Use the feature name, its description, and the list of options to guide your decision.\\
    - Select the ONE option that best matches the excerpt.\\
    - Do not summarise, explain, or add text.\\
    - Return only the numeric label of the chosen option.\newline \\
    
    \textbf{Context} \\
    Ritual: \{\texttt{ritual\_name}\} \\
    Feature: \{\texttt{feature\_name}\} \\
    Definition of feature: \{\texttt{feature\_description}\} \\
    Available options (numeric labels only): \{\texttt{feature\_options}\} \newline \\
    
    \textbf{Output Format} \\
    Return the numeric label of the selected option. \\
    No words, no punctuation, no extra output. \\
\end{quote}

\paragraph{Multi-task prompting prompt}
\begin{quote}
    \small 
    \raggedright
    \textbf{Instructions}\\
    You are given an ethnographic excerpt to classify across \{\texttt{n\_features}\} features in the following category: \{\texttt{category\_name}\}.\\
    - Carefully read the excerpt in full.\\
    - Use each feature name, its description, and the list of options to guide your decision.\\
    - For each feature, select the ONE option that best matches the excerpt.\\ 
    - Do not summarise, explain, or add text.\\
    - Return only the numeric label of the chosen option for each feature as a comma-separated list.\newline \\

    \textbf{Context}\\
    Ritual: \{\texttt{ritual\_name}\}\\
    Category: \{\texttt{category\_name}\}\\
    \{FOR \texttt{i} in \texttt{n\_features}\}\\
        \{\texttt{i}\}. \{feature\_name\}: \{feature\_description\}; \{feature\_options\}\\
    \{END FOR\}\newline \\

    \textbf{Output Format}\\
    Return a comma-separated list of exactly \{\texttt{n\_features}\} numeric values. Each value represents the chosen option for the corresponding feature in order (Example: "0,1,0,0,1,1"). 
\end{quote}

\paragraph{Perplexity baseline prompt}
\begin{quote}
    \small 
    \raggedright
    Based on the ritual "\{\texttt{ritual\_name}\}" from the \{\texttt{culture}\} culture (documented by \{\texttt{author}\} in \{\texttt{date}\} in "\{\texttt{ethnography\_title}\}"), does this ritual have the following feature?\newline \\

    Feature: \{\texttt{feature\_name}\}\\
    Definition: \{\texttt{feature\_description}\}\\
    Available options: \{\texttt{feature\_options}\}\newline \\

    Use your web search capabilities to find information about this ritual and culture. Return only the numeric label of the option that best matches. Do not include any explanation or additional text.
\end{quote}

\paragraph{Perplexity multi-task prompting prompt}
\begin{quote}
    \small 
    \raggedright
    Based on the ritual "\{\texttt{ritual\_name}\}" from the \{\texttt{culture}\} culture (documented by \{\texttt{author}\} in \{\texttt{date}\} in "\{\texttt{ethnography\_title}\}"), classify this ritual across \{\texttt{n\_features}\} features in the category "\{\texttt{category\_name}\}".\newline \\

    \{FOR \texttt{i} in \texttt{n\_features}\}\\
        \{\texttt{i}\}. \{feature\_name\}: \{feature\_description\}; \{feature\_options\}\\
    \{END FOR\} \newline \\

    Use your web search capabilities to find information about this ritual and culture. For each feature, select the ONE option that best matches. Return only a comma-separated list of exactly \{\texttt{n\_features}\} numeric values (Example: "0,1,0,0,1,1"). Do not include any explanation or additional text.
\end{quote}

\subsection*{Model performance}

\begin{figure*}[htbp]
    \centering
    \vspace{-7em}
    \includegraphics[width=1\linewidth]{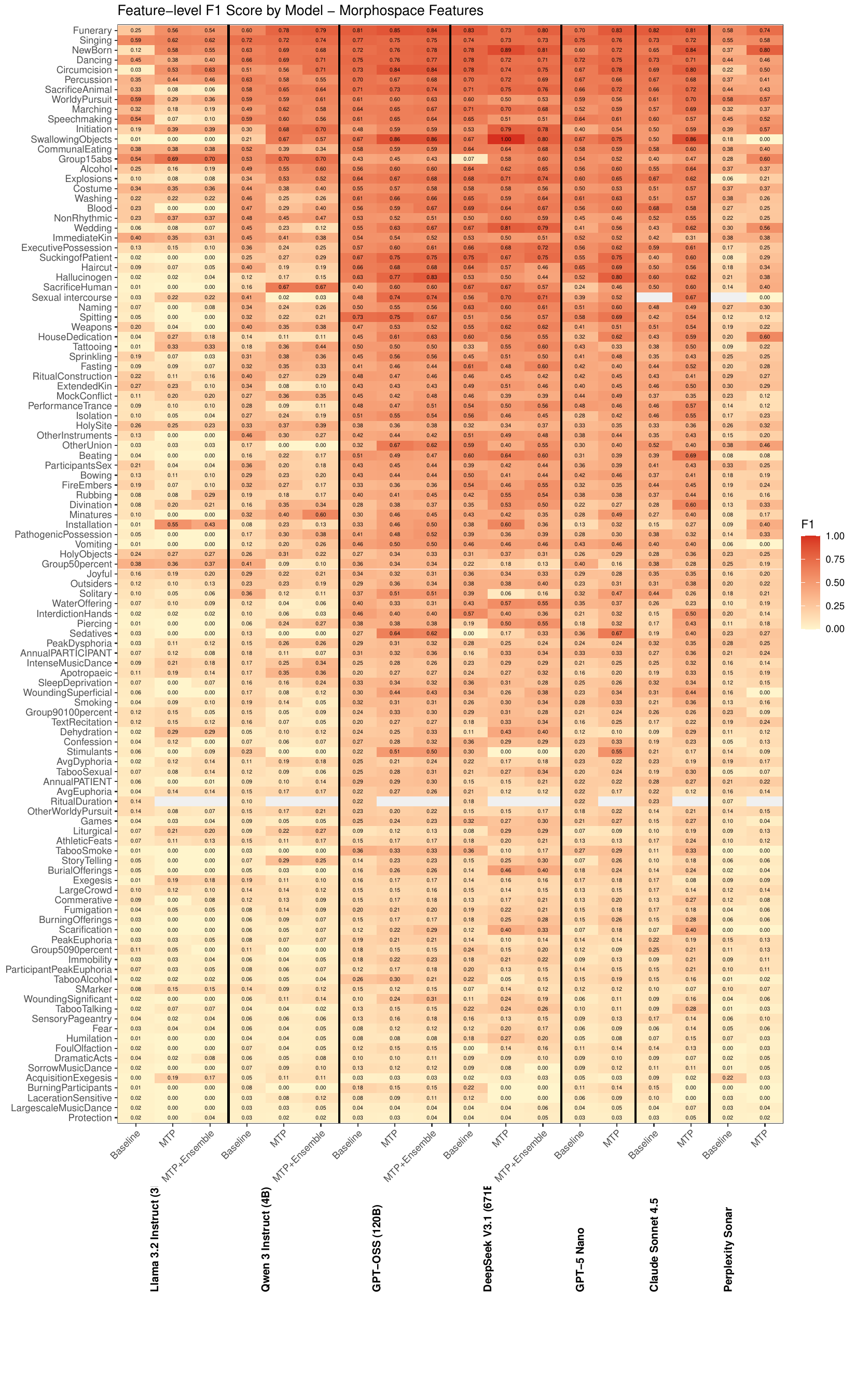}
    \vspace{-3em}
    \caption{\textbf{Feature-level F1 performance on all 115 ritual features.} RitualDuration was a single-feature category and thus only the baseline condition could be evaluated. Both Claude Sonnet 4.5 and Perplexity Sonar refused to answer questions pertaining to sexual intercourse due to model constraints.}
    \label{fig:feature-f1-heatmap-all}
\end{figure*}

Tables~\ref{tab:supp_all_model} and \ref{tab:supp_sync_model} report F1, precision, and recall scores for each model and prompting condition across the full feature set (115 features; see Figure~\ref{fig:feature-f1-heatmap-all}) and synchrony dataset (6 features), respectively. Table~\ref{tab:supp_sync_feature} provides feature-level performance for the synchrony dataset.

\begin{table*}[htbp]
\centering
\begin{tabular}{llccc}
\toprule
\textbf{Model} & \textbf{Condition} & \textbf{F1} & \textbf{Precision} & \textbf{Recall} \\
\midrule
\multirow{3}{*}{DeepSeek V3.1} & ensemble MTP & .415 & .358 & .640 \\
 & MTP & .407 & .372 & .621 \\
 & baseline & .382 & .313 & .680 \\
\midrule
\multirow{2}{*}{GPT-OSS 120B} & ensemble MTP & .411 & .339 & .718 \\
 & MTP & .406 & .332 & .715 \\
 & baseline & .362 & .281 & .746 \\
\midrule
\multirow{2}{*}{Claude Sonnet 4.5} & MTP & .394 & .328 & .658 \\
 & baseline & .326 & .249 & .764 \\
\midrule
\multirow{2}{*}{GPT-5 Nano} & MTP & .371 & .289 & .728 \\
 & baseline & .315 & .231 & .790 \\
\midrule
\multirow{3}{*}{Qwen 3 Instruct} & ensemble MTP & .254 & .205 & .648 \\
 & MTP & .249 & .197 & .649 \\
 & baseline & .238 & .169 & .779 \\
\midrule
\multirow{2}{*}{Perplexity Sonar} & MTP & .233 & .187 & .436 \\
 & baseline & .186 & .136 & .464 \\
\midrule
\multirow{3}{*}{Llama 3.2 Instruct} & MTP & .180 & .199 & .269 \\
 & ensemble MTP & .176 & .195 & .277 \\
 & baseline & .119 & .083 & .868 \\
\bottomrule
\end{tabular}
\caption{Model performance on the full feature set (115 features). F1, precision, and recall are macro-averaged across features.}
\label{tab:supp_all_model}
\end{table*}

\begin{table*}[htbp]
\centering
\begin{tabular}{llccc}
\toprule
\textbf{Model} & \textbf{Condition} & \textbf{F1} & \textbf{Precision} & \textbf{Recall} \\
\midrule
\multirow{2}{*}{GPT-OSS 120B} & MTP & .545 & .662 & .519 \\
 & ensemble MTP & .533 & .673 & .504 \\
 & baseline & .512 & .523 & .544 \\
\midrule
\multirow{2}{*}{GPT-5 Nano} & MTP & .538 & .569 & .561 \\
 & baseline & .491 & .473 & .625 \\
\midrule
\multirow{2}{*}{Claude Sonnet 4.5} & baseline & .512 & .435 & .744 \\
 & MTP & .450 & .564 & .451 \\
\midrule
\multirow{3}{*}{Qwen 3 Instruct} & MTP & .462 & .378 & .775 \\
 & ensemble MTP & .460 & .377 & .773 \\
 & baseline & .458 & .386 & .712 \\
\midrule
\multirow{2}{*}{Perplexity Sonar} & baseline & .337 & .301 & .419 \\
 & MTP & .295 & .306 & .320 \\
\midrule
\multirow{2}{*}{DeepSeek V3.1} & baseline & .335 & .618 & .263 \\
 & MTP & .291 & .470 & .295 \\
 & ensemble MTP & .310 & .471 & .299 \\
\midrule
\multirow{3}{*}{Llama 3.2 Instruct} & baseline & .334 & .210 & .987 \\
 & ensemble MTP & .257 & .297 & .435 \\
 & MTP & .251 & .292 & .425 \\
\bottomrule
\end{tabular}
\caption{Model performance on the synchrony dataset (6 features). F1, precision, and recall are macro-averaged across features.}
\label{tab:supp_sync_model}
\end{table*}

\begin{table*}[htbp]
\centering
\begin{tabular}{lllccc}
\toprule
\textbf{Feature} & \textbf{Model} & \textbf{Condition} & \textbf{F1} & \textbf{Prec} & \textbf{Rec} \\
\midrule
Sync. Dancing & GPT-OSS 120B & Ensemble MTP & \textbf{.809} & .905 & .731 \\
 & GPT-OSS 120B & MTP & .799 & .897 & .720 \\
 & Claude Sonnet 4.5 & MTP & .792 & .879 & .720 \\
\midrule
Sync. Singing & GPT-OSS 120B & Ensemble MTP & \textbf{.783} & .914 & .684 \\
 & GPT-OSS 120B & MTP & .779 & .914 & .679 \\
 & Claude Sonnet 4.5 & Baseline & .751 & .715 & .791 \\
\midrule
Sync. Chanting & GPT-5 Nano & MTP & \textbf{.598} & .583 & .615 \\
 & GPT-OSS 120B & MTP & .576 & .707 & .486 \\
 & GPT-OSS 120B & Ensemble MTP & .568 & .746 & .459 \\
\midrule
Sync. Praying & Claude Sonnet 4.5 & MTP & \textbf{.514} & .442 & .613 \\
 & GPT-5 Nano & MTP & .457 & .425 & .493 \\
 & Qwen 3 Instruct & MTP & .404 & .266 & .840 \\
\midrule
Sync. Marching & GPT-OSS 120B & Baseline & \textbf{.427} & .297 & .760 \\
 & GPT-OSS 120B & Ensemble MTP & .409 & .286 & .720 \\
 & GPT-OSS 120B & MTP & .409 & .286 & .720 \\
\midrule
Sync. Generic Mvmt & GPT-5 Nano & Baseline & \textbf{.461} & .425 & .504 \\
 & Claude Sonnet 4.5 & Baseline & .439 & .310 & .752 \\
 & Qwen 3 Instruct & Ensemble MTP & .415 & .289 & .736 \\
\bottomrule
\end{tabular}
\caption{Feature-level performance on the synchrony dataset. Best F1 per feature shown in bold.}
\label{tab:supp_sync_feature}
\end{table*}

For the full feature set, all models outperformed the majority-class baseline on 98.2\% of features (109 of 111 evaluable features). The mean majority-class baseline F1 was .019, compared to a mean best-model F1 of .472, representing a mean improvement of .453 F1 points. For the synchrony dataset, all models outperformed the baseline on all 6 features.

\subsection*{Inter-coder reliability and agreement analysis}

For the synchrony dataset, two independent human coders annotated each ritual, with disagreements resolved through discussion. Table~\ref{tab:supp_human_reliability} reports inter-coder reliability metrics for each synchrony feature. Only one feature (synchronous singing) achieved substantial agreement ($\kappa \geq 0.80$); two features (synchronous marching and synchronous generic movement) exhibited poor agreement ($\kappa < 0.40$). McNemar's tests revealed significant systematic bias between coders for five of six features (all except synchronous marching), with Coder 1 generally detecting more instances than Coder 2.

\begin{table*}[htbp]
\centering
\begin{tabular}{lccccc}
\toprule
\textbf{Feature} & \textbf{N} & \textbf{Raw Agr.} & \textbf{$\kappa$} & \textbf{PA} & \textbf{NA} \\
\midrule
Sync. Singing & 606 & .965 & .920 & .945 & .975 \\
Sync. Dancing & 585 & .909 & .792 & .858 & .934 \\
Sync. Praying & 591 & .932 & .647 & .683 & .962 \\
Sync. Chanting & 598 & .841 & .464 & .558 & .903 \\
Sync. Marching & 600 & .935 & .367 & .400 & .966 \\
Sync. Generic Mvmt & 588 & .730 & .248 & .404 & .825 \\
\midrule
\textbf{Mean} & --- & .885 & .573 & .641 & .927 \\
\bottomrule
\end{tabular}
\caption{Human inter-coder reliability for synchrony features. $\kappa$ = Cohen's kappa; PA = positive agreement; NA = negative agreement.}
\label{tab:supp_human_reliability}
\end{table*}

Table~\ref{tab:supp_llm_f1_by_feature} reports LLM F1 scores for each synchrony feature by model (baseline condition), alongside human reliability. Human inter-coder agreement correlated positively with LLM performance across features ($r = .61$). The single feature with near-perfect human agreement (synchronous singing, $\kappa = .92$) yielded the highest mean LLM F1 (.57). Features with poor human agreement ($\kappa < .40$) yielded substantially lower mean LLM F1 (.31).

\begin{table*}[htbp]
\centering
\begin{tabular}{lcccccccc}
\toprule
\textbf{Feature} & \textbf{$\kappa$} & \textbf{Claude} & \textbf{DeepS.} & \textbf{GPT-5} & \textbf{GPT-OSS} & \textbf{Llama} & \textbf{Perpl.} & \textbf{Qwen} \\
\midrule
Sync. Singing & .92 & .751 & .306 & .622 & .748 & .492 & .475 & .621 \\
Sync. Dancing & .79 & .791 & .516 & .651 & .757 & .483 & .463 & .739 \\
Sync. Chanting & .46 & .442 & .414 & .500 & .434 & .332 & .296 & .419 \\
Sync. Praying & .65 & .451 & .244 & .407 & .331 & .239 & .292 & .349 \\
Sync. Marching & .37 & .196 & .361 & .305 & .427 & .086 & .189 & .221 \\
Sync. Gen. Mvmt & .25 & .439 & .170 & .461 & .372 & .373 & .309 & .398 \\
\bottomrule
\end{tabular}
\caption{LLM F1 scores by synchrony feature (baseline condition). Human $\kappa$ shown for reference.}
\label{tab:supp_llm_f1_by_feature}
\end{table*}

\begin{figure*}[htbp]
    \centering
    \includegraphics[width=1\linewidth]{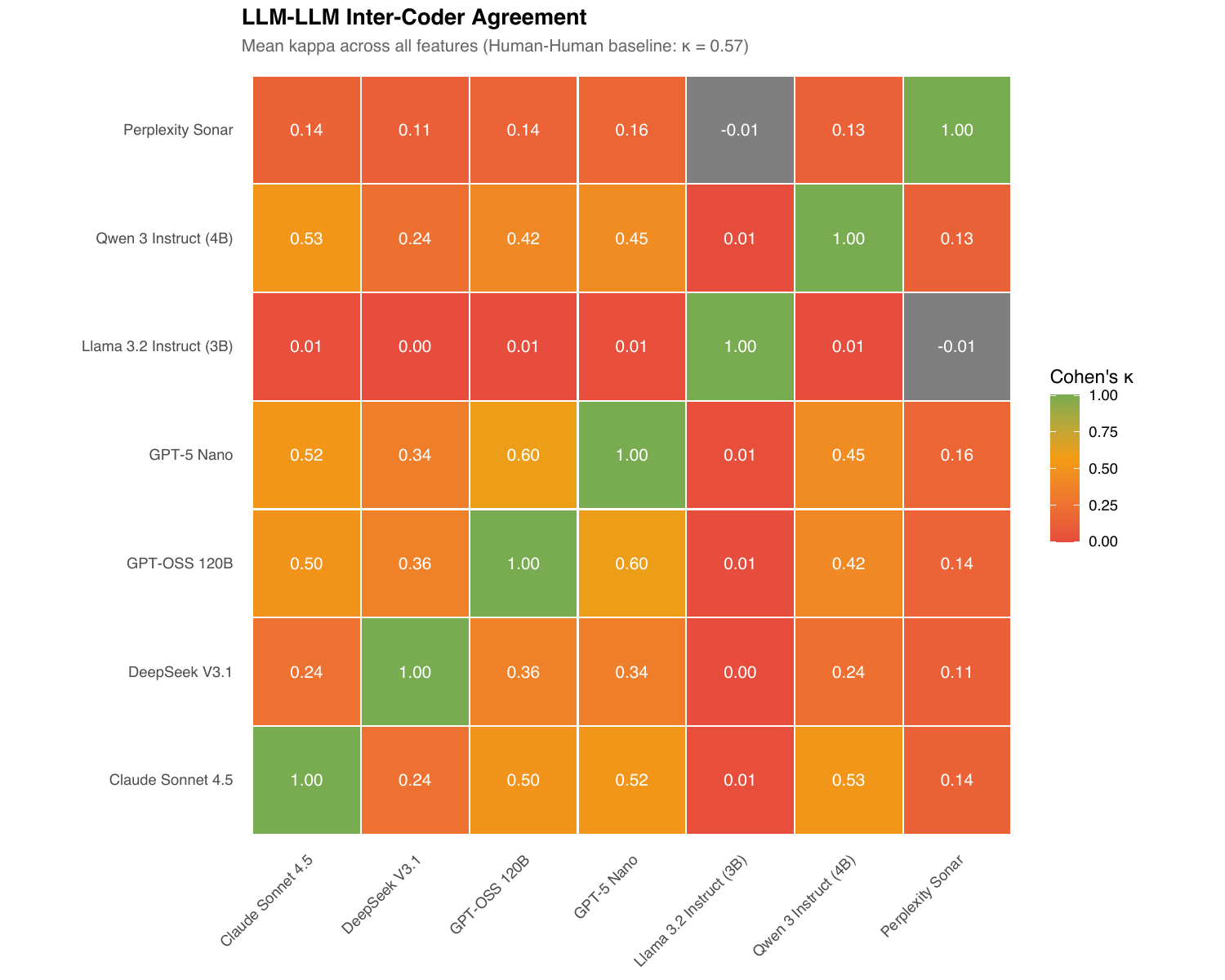}
    \caption{LLM–LLM agreement heatmap. Mean Cohen's $\kappa$ is reported across all model pairs.}
    \label{fig:supp_llm_llm_fig}
\end{figure*}

To examine whether LLMs converge on similar annotations despite individual errors, we computed pairwise Cohen's $\kappa$ between all LLM pairs (21 pairs across 7 models) for each synchrony feature. Table~\ref{tab:supp_llm_llm_summary} and Figure~\ref{fig:supp_llm_llm_fig} summarises LLM-LLM agreement compared to human-human agreement.

LLMs agreed with each other substantially less than human coders agreed with each other. Mean LLM-LLM $\kappa$ was .233 compared to human-human $\kappa$ of .573, a difference of .340. For all six features, LLM-LLM agreement fell below human-human agreement. The gap was largest for features that humans found easiest: synchronous singing (LLM-LLM $\kappa$=.271 vs.\ human $\kappa$=.920, diff=.649) and synchronous dancing (LLM-LLM $\kappa$=.346 vs.\ human $\kappa$=.792, diff=.446). For features with poor human agreement, the gap was smaller but still present: synchronous generic movement (LLM-LLM $\kappa$=.149 vs.\ human $\kappa$=.248, diff=.099).

\begin{table*}[htbp]
\centering
\begin{tabular}{lccccc}
\toprule
\textbf{Feature} & \textbf{LLM-LLM $\kappa$} & \textbf{Human $\kappa$} & \textbf{Diff} & \textbf{Min} & \textbf{Max} \\
\midrule
Sync. Singing & .271 & .920 & $-$.649 & $-$.003 & .710 \\
Sync. Dancing & .346 & .792 & $-$.446 & $-$.026 & .773 \\
Sync. Praying & .162 & .647 & $-$.485 & $-$.005 & .534 \\
Sync. Chanting & .230 & .464 & $-$.234 & $-$.009 & .608 \\
Sync. Marching & .239 & .367 & $-$.128 & $-$.001 & .589 \\
Sync. Generic Mvmt & .149 & .248 & $-$.099 & $-$.001 & .500 \\
\midrule
\textbf{Mean} & .233 & .573 & $-$.340 & --- & --- \\
\bottomrule
\end{tabular}
\caption{LLM-LLM agreement compared to human-human agreement. LLM-LLM $\kappa$ is the mean across all 21 model pairs; Min and Max show the range of pairwise agreement.}
\label{tab:supp_llm_llm_summary}
\end{table*}

Table~\ref{tab:supp_llm_llm_pairs} reports mean pairwise agreement across features for each model pair. Models from similar architectures or parameter scales tended to agree more strongly. The highest agreement was between GPT-OSS 120B and GPT-5 Nano ($\kappa$=.600), followed by Claude Sonnet 4.5 paired with Qwen 3 Instruct ($\kappa$=.526), GPT-5 Nano ($\kappa$=.522), and GPT-OSS 120B ($\kappa$=.502). In contrast, Llama 3.2 Instruct showed near-zero agreement with all other models ($\kappa<.012$), consistent with its tendency to predict presence in nearly all cases. Perplexity Sonar exhibited low agreement with other models ($\kappa$=.109–.156), likely reflecting its distinct retrieval-augmented architecture.

\begin{table*}[htbp]
\centering
\begin{tabular}{llccc}
\toprule
\textbf{Model 1} & \textbf{Model 2} & \textbf{Mean $\kappa$} & \textbf{Min} & \textbf{Max} \\
\midrule
GPT-OSS 120B & GPT-5 Nano & .600 & .500 & .686 \\
Claude Sonnet 4.5 & Qwen 3 Instruct & .526 & .338 & .773 \\
Claude Sonnet 4.5 & GPT-5 Nano & .522 & .345 & .644 \\
Claude Sonnet 4.5 & GPT-OSS 120B & .502 & .222 & .771 \\
GPT-5 Nano & Qwen 3 Instruct & .447 & .270 & .558 \\
GPT-OSS 120B & Qwen 3 Instruct & .416 & .165 & .662 \\
DeepSeek V3.1 & GPT-OSS 120B & .362 & .158 & .534 \\
DeepSeek V3.1 & GPT-5 Nano & .336 & .111 & .586 \\
Claude Sonnet 4.5 & DeepSeek V3.1 & .238 & .102 & .449 \\
DeepSeek V3.1 & Qwen 3 Instruct & .235 & .060 & .430 \\
\midrule
\multicolumn{5}{l}{\textit{Lowest agreement pairs}} \\
Claude Sonnet 4.5 & Llama 3.2 Instruct & .011 & .005 & .018 \\
GPT-5 Nano & Llama 3.2 Instruct & .010 & .004 & .017 \\
GPT-OSS 120B & Llama 3.2 Instruct & .006 & .002 & .011 \\
Llama 3.2 Instruct & Perplexity Sonar & $-$.006 & $-$.026 & $-$.001 \\
\bottomrule
\end{tabular}
\caption{Pairwise LLM-LLM agreement (mean $\kappa$ across 6 synchrony features). Top 10 and bottom 4 pairs shown.}
\label{tab:supp_llm_llm_pairs}
\end{table*}

These results indicate that different LLMs interpret ambiguous ethnographic descriptions in systematically different ways. Even when individual models achieve similar overall accuracy, they often disagree on specific cases, limiting the potential for simple ensemble or majority-voting approaches to improve reliability.

\subsection*{Mixed-effects regression models}

To identify factors influencing LLM performance, we fitted nested mixed-effects models at two levels of analysis: ritual-level (using F1 score as the outcome) and prediction-level (using binary correctness for each feature annotation). At the prediction level, we stratified by true class (positive vs negative cases) to separately examine detection of present features and correct rejection of absent features.

\paragraph{Ritual-level analysis}

Table~\ref{tab:supp_ritual_r2} reports variance explained (R²) by nested models at the ritual level. Model factors (model identity and prompting condition) explained 33.2\% of variance. Adding text factors (character length, readability, lexical diversity, region, publication year) increased this to 36.0\%. Task factors (proportion binary features, feature description length) contributed minimally, reaching 36.6\%. The conditional R² (including random intercepts for ritual) was 64.9\%, indicating substantial unexplained between-ritual variation.

\begin{table*}[htbp]
\centering
\begin{tabular}{lccc}
\toprule
\textbf{Model} & \textbf{Predictors} & \textbf{Marginal $R^2$} & \textbf{Conditional $R^2$} \\
\midrule
M1 & Model factors & .332 & .645 \\
M2 & + Text factors & .360 & .649 \\
M3 & + Task factors & .366 & .649 \\
\bottomrule
\end{tabular}
\caption{Variance explained by nested models at the ritual level.}
\label{tab:supp_ritual_r2}
\end{table*}

Table~\ref{tab:supp_ritual_coefs} reports key fixed effect coefficients from the final model (M3). Text length was negatively associated with performance ($\beta$=-0.019, $p<$.001). North America and Oceania showed significantly lower F1 scores than the reference region ($p<$.02). Longer feature descriptions predicted poorer performance ($\beta$=-0.020, $p$=.001). Readability (Gunning FOG) and lexical diversity (MTLD) were not significant predictors.

\begin{table*}[htbp]
\centering
\begin{tabular}{lcccc}
\toprule
\textbf{Predictor} & \textbf{$\beta$} & \textbf{SE} & \textbf{$t$} & \textbf{$p$} \\
\midrule
\multicolumn{5}{l}{\textit{Model factors (vs GPT-OSS 120B)}} \\
DeepSeek V3.1 & 0.007 & 0.005 & 1.21 & .226 \\
GPT-5 Nano & $-$0.047 & 0.005 & $-$8.69 & $<$.001 \\
Llama 3.2 & $-$0.242 & 0.005 & $-$44.77 & $<$.001 \\
Claude Sonnet 4.5 & $-$0.025 & 0.005 & $-$4.62 & $<$.001 \\
Qwen 3 & $-$0.098 & 0.005 & $-$18.21 & $<$.001 \\
\midrule
\multicolumn{5}{l}{\textit{Condition (vs baseline)}} \\
MTP & 0.036 & 0.005 & 6.63 & $<$.001 \\
Ensemble MTP & 0.038 & 0.005 & 7.01 & $<$.001 \\
\midrule
\multicolumn{5}{l}{\textit{Text factors}} \\
Text length (log, z) & $-$0.019 & 0.004 & $-$5.35 & $<$.001 \\
Gunning FOG (z) & $-$0.005 & 0.004 & $-$1.29 & .197 \\
MTLD (z) & $-$0.002 & 0.004 & $-$0.52 & .602 \\
Region: N. America & $-$0.036 & 0.015 & $-$2.37 & .018 \\
Region: Oceania & $-$0.035 & 0.014 & $-$2.58 & .010 \\
Publication year (z) & $-$0.002 & 0.004 & $-$0.56 & .573 \\
\midrule
\multicolumn{5}{l}{\textit{Task factors}} \\
Prop. binary features & $-$1.949 & 0.692 & $-$2.82 & .005 \\
Feature desc. length (z) & $-$0.020 & 0.006 & $-$3.28 & .001 \\
\bottomrule
\end{tabular}
\caption{Key fixed effect coefficients from ritual-level model M3.}
\label{tab:supp_ritual_coefs}
\end{table*}

Variance decomposition showed that 44.6\% of residual variance lay between rituals (random intercept variance = 0.0063), with 55.4\% within rituals across model-condition combinations (residual variance = 0.0078).

\paragraph{Prediction-level analysis (stratified)}

At the prediction level, we analysed over 1.1 million individual feature annotations, stratified into positive cases (feature present in ground truth; 16.0\% of predictions) and negative cases (feature absent; 84.0\%). Table~\ref{tab:supp_pred_r2} reports variance explained by stratum.

\begin{table*}[htbp]
\centering
\begin{tabular}{llccc}
\toprule
\textbf{Stratum} & \textbf{Model} & \textbf{Predictors} & \textbf{Marg. $R^2$} & \textbf{Cond. $R^2$} \\
\midrule
\multirow{3}{*}{Positive} & M1 & Model factors & .036 & .299 \\
 & M2 & + Text factors & .038 & .299 \\
 & M3 & + Task factors & .225 & .373 \\
\midrule
\multirow{3}{*}{Negative} & M1 & Model factors & .249 & .552 \\
 & M2 & + Text factors & .293 & .552 \\
 & M3 & + Task factors & .348 & .524 \\
\bottomrule
\end{tabular}
\caption{Variance explained by nested models at the prediction level (stratified).}
\label{tab:supp_pred_r2}
\end{table*}

For positive cases (detection), task factors dramatically increased marginal R² from 3.8\% to 22.5\%, driven primarily by feature type: multiclass features were 90\% less likely to be correctly detected than binary features (OR=0.10, $p<$.001). For negative cases (specificity), model and text factors explained more variance (34.8\% total), with feature base rate as the strongest task predictor: rarer features were more likely to be correctly rejected as absent (OR=0.23 per SD increase in base rate, $p<$.001).

Table~\ref{tab:supp_pred_or} reports odds ratios for key task predictors by stratum.

\begin{table*}[htbp]
\centering
\begin{tabular}{llcccc}
\toprule
\textbf{Predictor} & \textbf{Stratum} & \textbf{OR} & \textbf{95\% CI} & \textbf{$p$} \\
\midrule
\multirow{2}{*}{Multiclass (vs binary)} & Positive & 0.10 & [0.03, 0.35] & $<$.001 \\
 & Negative & 3.99 & [1.39, 11.44] & .010 \\
\midrule
\multirow{2}{*}{Feature desc. length (z)} & Positive & 1.02 & [0.88, 1.19] & .775 \\
 & Negative & 0.85 & [0.64, 1.13] & .265 \\
\midrule
\multirow{2}{*}{Feature base rate (z)} & Positive & 1.09 & [0.75, 1.59] & .636 \\
 & Negative & 0.23 & [0.16, 0.35] & $<$.001 \\
\bottomrule
\end{tabular}
\caption{Odds ratios for task predictors by stratum (prediction-level M3).}
\label{tab:supp_pred_or}
\end{table*}

Variance decomposition at the prediction level revealed that between-feature variation dominated in both strata. For positive cases, 87.4\% of random effect variance was between features (variance = 0.68) versus 12.6\% between rituals (variance = 0.10). For negative cases, the pattern was similar: 86.7\% between features (variance = 1.05) versus 13.3\% between rituals (variance = 0.16). This indicates that feature characteristics, rather than text characteristics, are the primary source of unexplained variation in LLM annotation accuracy.

\subsection*{Regional analysis of model performance}

\begin{figure*}[htbp]
    \centering
    \includegraphics[width=1\linewidth]{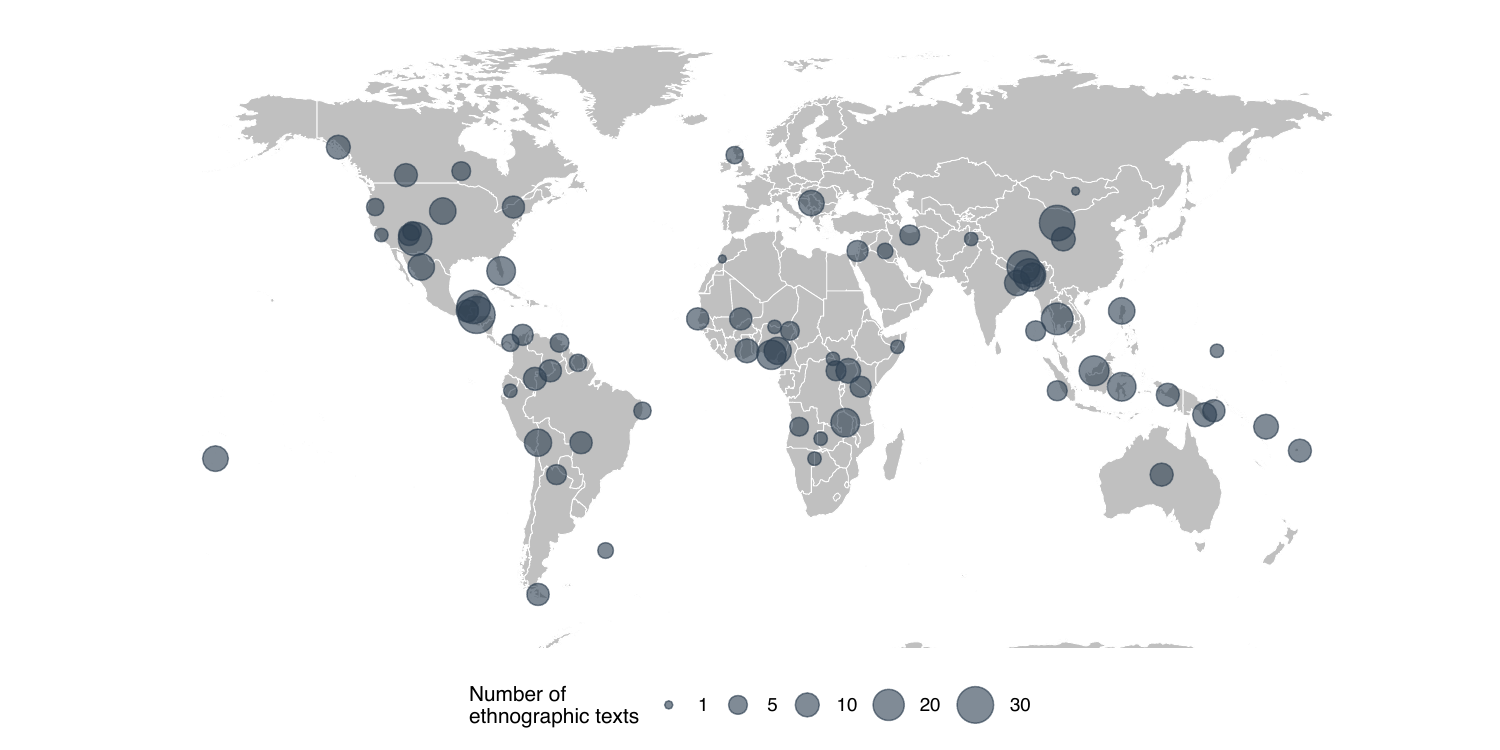}
    \caption{Ethnographic text coverage across the world.}
    \label{fig:map-coverage}
\end{figure*}

To assess whether LLM performance varied by the geographic origin of ethnographic texts, we conducted Kruskal-Wallis tests for each model separately, comparing F1 scores across 9 world regions. Table~\ref{tab:region_sample} reports the distribution of texts across regions; Table~\ref{tab:region_kruskal} summarises omnibus test results; Table~\ref{tab:region_pairwise} lists significant pairwise comparisons (Bonferroni-corrected Wilcoxon tests). Figure \ref{fig:map-coverage} shows the coverage of ethnographic texts across the world.

\begin{table*}[htbp]
\centering
\begin{tabular}{lcc}
\toprule
\textbf{Region} & \textbf{N Texts} & \textbf{\%} \\
\midrule
Asia & 155 & 29.0 \\
Africa & 93 & 17.4 \\
M. America \& Caribbean & 70 & 13.1 \\
Oceania & 66 & 12.3 \\
S. America & 62 & 11.6 \\
N. America & 47 & 8.8 \\
Europe & 23 & 4.3 \\
Middle East & 13 & 2.4 \\
Australia & 6 & 1.1 \\
\midrule
\textbf{Total} & \textbf{535} & \textbf{100.0} \\
\bottomrule
\end{tabular}
\caption{Distribution of ethnographic texts by geographic region.}
\label{tab:region_sample}
\end{table*}

\begin{table*}[htbp]
\centering
\begin{tabular}{lccc}
\toprule
\textbf{Model} & \textbf{$\chi^2$} & \textbf{df} & \textbf{$p$} \\
\midrule
GPT-OSS 120B & 41.34 & 8 & $<$.0001 \\
Perplexity Sonar & 36.40 & 8 & $<$.0001 \\
Qwen 3 Instruct & 34.58 & 8 & $<$.0001 \\
GPT-5 Nano & 31.23 & 8 & .0001 \\
Claude Sonnet 4.5 & 26.44 & 8 & .0009 \\
DeepSeek V3.1 & 25.76 & 8 & .0012 \\
Llama 3.2 Instruct & 11.35 & 8 & .18 \\
\bottomrule
\end{tabular}
\caption{Kruskal-Wallis tests for regional differences in F1 score by model.}
\label{tab:region_kruskal}
\end{table*}

\begin{table*}[htbp]
\centering
\begin{tabular}{llc}
\toprule
\textbf{Model} & \textbf{Comparison} & \textbf{$p$} \\
\midrule
\multirow{6}{*}{GPT-OSS 120B} & N. America $<$ Africa & $<$.0001 \\
 & Asia $<$ Africa & .003 \\
 & Oceania $<$ Africa & .015 \\
 & N. America $<$ M. America \& Carib & .039 \\
 & N. America $<$ Middle East & .012 \\
 & N. America $<$ S. America & .028 \\
\midrule
\multirow{6}{*}{Perplexity Sonar} & Africa $<$ Europe & $<$.001 \\
 & Asia $<$ Europe & $<$.001 \\
 & M. America \& Carib $<$ Europe & .002 \\
 & N. America $<$ Europe & .019 \\
 & Oceania $<$ Europe & $<$.0001 \\
 & S. America $<$ Europe & $<$.0001 \\
\midrule
\multirow{5}{*}{GPT-5 Nano} & N. America $<$ Africa & $<$.001 \\
 & N. America $<$ Asia & .010 \\
 & N. America $<$ M. America \& Carib & .003 \\
 & N. America $<$ Middle East & .009 \\
 & N. America $<$ S. America & .037 \\
\midrule
\multirow{3}{*}{Qwen 3 Instruct} & N. America $<$ M. America \& Carib & $<$.001 \\
 & Oceania $<$ M. America \& Carib & .013 \\
 & N. America $<$ Middle East & .025 \\
\midrule
\multirow{2}{*}{Claude Sonnet 4.5} & Oceania $<$ Africa & .006 \\
 & Oceania $<$ Europe & .025 \\
\midrule
DeepSeek V3.1 & N. America $<$ Africa & .036 \\
\bottomrule
\end{tabular}
\caption{Significant pairwise regional comparisons (Bonferroni-corrected $p<$.05). Comparisons show the first region performing worse than the second.}
\label{tab:region_pairwise}
\end{table*}

Six of seven models showed significant omnibus effects of region. North America consistently exhibited lower F1 scores across GPT-OSS 120B, GPT-5 Nano, Qwen 3, and DeepSeek V3.1; Oceania showed lower performance for GPT-OSS 120B, Qwen 3, and Claude Sonnet 4.5. Perplexity Sonar displayed a distinct pattern: Europe significantly outperformed all other regions, possibly reflecting its web-augmented architecture drawing on online sources that may be more comprehensive for European ethnographic contexts. Only Llama 3.2 Instruct showed no significant regional variation. The causes of these regional differences remain unclear but may relate to variation in ethnographic writing conventions, ritual complexity, or representation in model training data.

\subsection*{Ensemble certainty and annotation accuracy}

\begin{figure*}[htbp]
    \centering
    \includegraphics[width=1\linewidth]{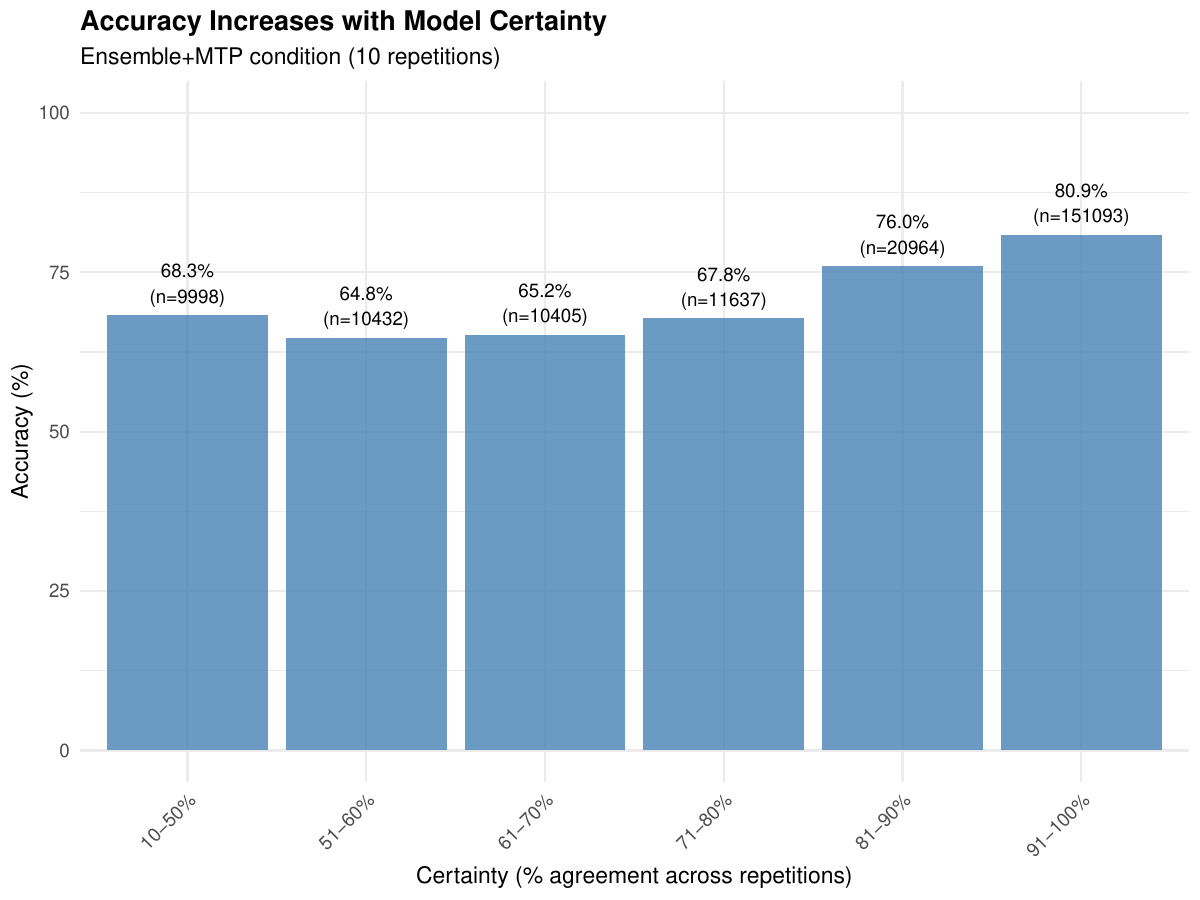}
    \caption{Model certainty is a good predictor of annotation accuracy.}
    \label{fig:accuracy-by-certainty}
\end{figure*}

For the ensemble+MTP condition, we examined whether model certainty (the percentage of 10 repetitions agreeing with the modal annotation) predicted correctness. Table~\ref{tab:supp_certainty_model} reports correlations by model; Table~\ref{tab:supp_certainty_bins} shows accuracy by certainty bin; Figure \ref{fig:accuracy-by-certainty} demonstrates the relationships between certainty and accuracy.

\begin{table*}[htbp]
\centering
\begin{tabular}{lcccc}
\toprule
\textbf{Model} & \textbf{N} & \textbf{Mean Certainty} & \textbf{Accuracy} & \textbf{$r$} \\
\midrule
GPT-OSS 120B & 64,798 & 96.2\% & 82.2\% & .360 \\
DeepSeek V3.1 & 52,585 & 91.3\% & 83.8\% & .230 \\
Qwen 3 Instruct & 32,348 & 99.5\% & 66.6\% & .076 \\
Llama 3.2 Instruct & 64,798 & 84.4\% & 73.4\% & $-$.015 \\
\bottomrule
\end{tabular}
\caption{Correlation between ensemble certainty and annotation correctness by model.}
\label{tab:supp_certainty_model}
\end{table*}

\begin{table*}[htbp]
\centering
\begin{tabular}{lccc}
\toprule
\textbf{Certainty Bin} & \textbf{N} & \textbf{\% of Predictions} & \textbf{Accuracy} \\
\midrule
10--50\% & 9,998 & 4.7\% & 68.3\% \\
51--60\% & 10,432 & 4.9\% & 64.8\% \\
61--70\% & 10,405 & 4.9\% & 65.2\% \\
71--80\% & 11,637 & 5.4\% & 67.8\% \\
81--90\% & 20,964 & 9.8\% & 76.0\% \\
91--100\% & 151,093 & 70.4\% & 80.9\% \\
\bottomrule
\end{tabular}
\caption{Annotation accuracy by certainty bin (ensemble+MTP condition, all models pooled).}
\label{tab:supp_certainty_bins}
\end{table*}

The overall correlation between certainty and correctness was modest but statistically significant ($r$=0.12, 95\% CI: 0.12--0.12, $p<$.001; $N$=214,529). A logistic regression confirmed that each 10 percentage point increase in certainty was associated with 18\% higher odds of a correct annotation (OR=1.18). The relationship was driven primarily by the better-performing models (GPT-OSS 120B and DeepSeek V3.1), suggesting that ensemble consistency is a more reliable quality indicator for larger models.
\newpage
\bibliographystyle{unsrt}  
\bibliography{References}
\end{document}